\documentclass[]{wan}

\usepackage{wrapfig}
\usepackage{tabularx}
\usepackage{textcomp}
\usepackage{stfloats}
\usepackage{url}
\usepackage{verbatim}
\usepackage{graphicx}
\usepackage{titlesec}
\usepackage{tocloft}
\usepackage{adjustbox}
\usepackage{multirow}
\usepackage{pifont}
\usepackage{tikz}
\usepackage{comment}
\usepackage{amsmath,amssymb} 
\usepackage{colortbl}  
\usepackage{color}
\usepackage{booktabs} 
\usepackage{hyperref}
\usepackage{graphicx}   
\usepackage{subcaption} 
\usepackage{booktabs}
\usepackage{amsmath} %
\usepackage{multirow} %
\usepackage{booktabs} %
\usepackage{subcaption} %
\usepackage{graphicx}   %
\usepackage{wrapfig}    %
\usepackage{caption}    %

\RequirePackage{xspace}
\makeatletter
\DeclareRobustCommand\onedot{\futurelet\@let@token\@onedot}
\def\@onedot{\ifx\@let@token.\else.\null\fi\xspace}

\def\eg{\emph{e.g}\onedot}

\def\etc{\emph{etc}\onedot}

\makeatother

\usepackage{makecell}
\definecolor{Gray}{gray}{0.95}

\definecolor{adptorange}{RGB}{248, 205, 172}
\definecolor{cmpblue}{RGB}{189, 215, 238}
\definecolor{cmpblue}{RGB}{189, 215, 238}

\definecolor{our_red}{RGB}{232,157,160}
\definecolor{our_blue}{RGB}{136,206,230}
\definecolor{our_orange}{RGB}{246,200,168}
\definecolor{our_green}{RGB}{178,211,164}

\definecolor{attn_code0}{RGB}{247,215,200}
\definecolor{attn_code1}{RGB}{238,169,139}
\definecolor{mlp_code0}{RGB}{204,201,221}
\definecolor{mlp_code1}{RGB}{102,95,153}

\definecolor{token_blue}{RGB}{84, 120, 140}

\definecolor{myMagenta}{rgb}{0.9,0,0.4}

\usepackage{pifont}       
\usepackage{bbding}       
\usepackage{fontawesome}
\usepackage{xspace}

\usepackage{float}

\newlength\savewidth

\newcolumntype{x}[1]{>{\centering\arraybackslash}p{#1pt}}
\newcolumntype{y}[1]{>{\raggedright\arraybackslash}p{#1pt}}
\newcolumntype{z}[1]{>{\raggedleft\arraybackslash}p{#1pt}}

\usepackage{colortbl}
\usepackage{xcolor}
\usepackage{wrapfig}

\renewcommand{\paragraph}[1]{\vspace{1.25mm}\noindent\textbf{#1}}

\usepackage{algorithm}
\usepackage{listings}

\definecolor{codeblue}{rgb}{0.25, 0.5, 0.5}
\definecolor{codekw}{rgb}{0.35, 0.35, 0.75}
\lstdefinestyle{Pytorch}{
    language = Python,
    backgroundcolor = \color{white},
    basicstyle = \fontsize{9pt}{8pt}\selectfont\ttfamily\bfseries,
    columns = fullflexible,
    aboveskip=1pt,
    belowskip=1pt,
    breaklines = true,
    captionpos = b,
    commentstyle = \color{codeblue},
    keywordstyle = \color{codekw},
}

\definecolor{green}{HTML}{009000}
\definecolor{red}{HTML}{ea4335}

\newcommand{\wan}{{Wan-Image}\xspace}
\title{Wan-Image: Pushing the Boundaries of Generative Visual Intelligence}

\author{Wan Team, Alibaba Group}

\abstract{

We present Wan-Image, a unified visual generation system explicitly engineered to paradigm-shift image generation models from casual synthesizers into professional-grade productivity tools. While contemporary diffusion models excel at aesthetic generation, they frequently encounter critical bottlenecks in rigorous design workflows that demand absolute controllability, complex typography rendering, and strict identity preservation. To address these challenges, Wan-Image features a natively unified multi-modal architecture by synergizing the cognitive capabilities of large language models with the high-fidelity pixel synthesis of diffusion transformers, which seamlessly translates highly nuanced user intents into precise visual outputs. 
It is fundamentally powered by large-scale multi-modal data scaling, a systematic fine-grained annotation engine, and curated reinforcement learning data to surpass basic instruction following and unlock expert-level professional capabilities.
These include ultra-long complex text rendering, hyper-diverse portrait generation, palette-guided generation, multi-subject identity preservation, coherent sequential visual generation, precise multi-modal interactive editing, native alpha-channel generation, and high-efficiency 4K synthesis. Across diverse human evaluations, Wan-Image exceeds Seedream 5.0 Lite and GPT Image 1.5 in overall performance, reaching parity with Nano Banana Pro in challenging tasks. Ultimately, Wan-Image revolutionizes visual content creation across e-commerce, entertainment, education, and personal productivity, redefining the boundaries of professional visual synthesis.

}

\date{\today} 
\modelcard{Wan2.7-Image}
\webservice{\href{https://create.wan.video/generate/image/generate?model=wan2.7-pro}{create.wan.video/generate/image/generate?model=wan2.7-pro}}

\begin{document}
\thispagestyle{firstheader}
\maketitle
\pagestyle{empty}

    

\begin{figure}[!hb]
    \centering
    \vspace{-2mm}
    \makebox[\linewidth][c]{%
    \includegraphics[width=0.78\linewidth]{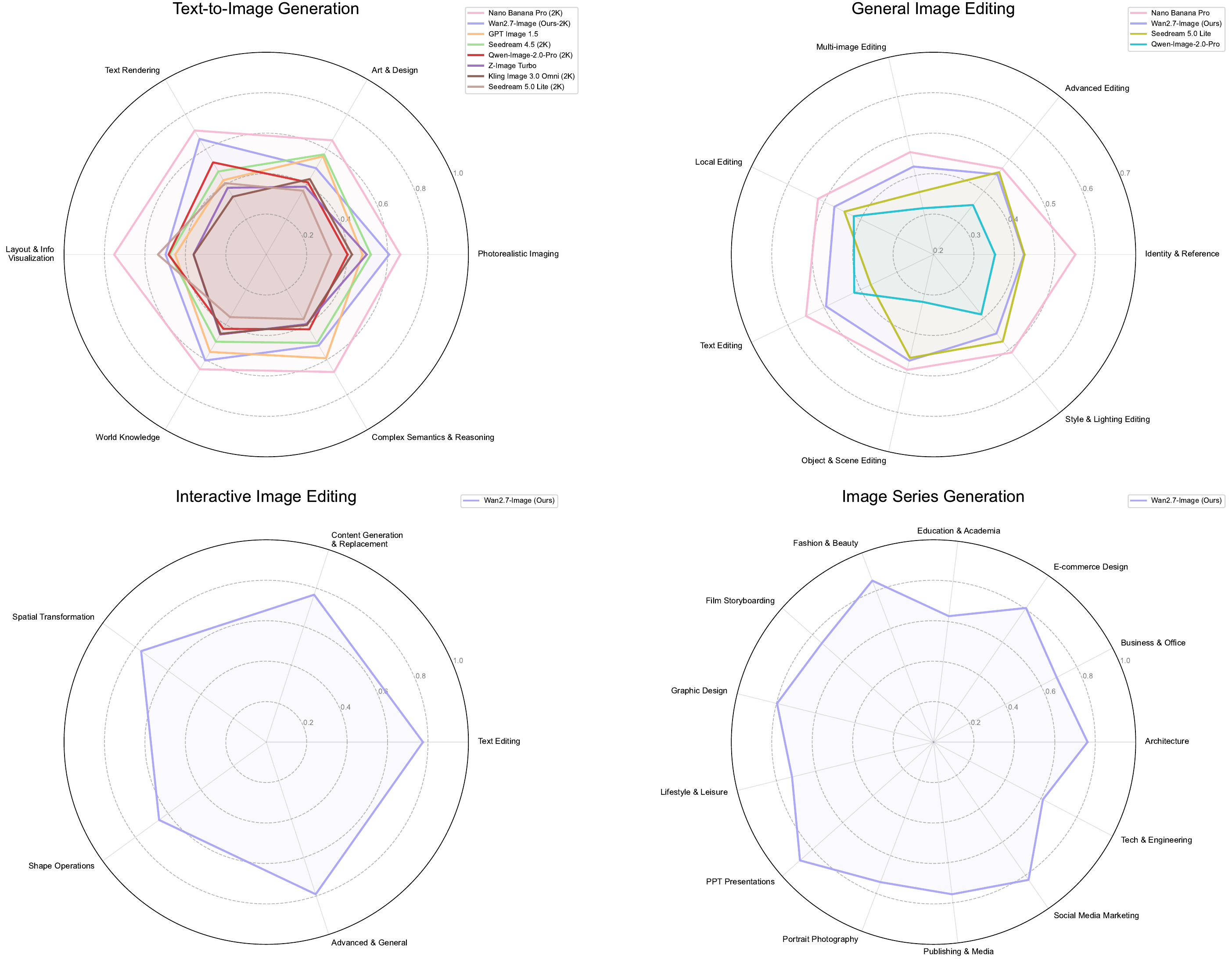}%
  }
    
    \caption{Comprehensive Human Evaluation across diverse generative tasks. We report a composite \textbf{Win Score} ($\frac{\text{Wins} + 0.5 \times \text{BothGood}}{\text{Total}}$) for Text-to-Image and General Editing, and \textbf{Pass Rate} for Interactive Editing and Series Generation.}
    \vspace{-20mm}
    \label{fig:radar_report}
\end{figure}

\begin{figure}[!htb]
    \centering
    \vspace{-20pt}
    \makebox[\linewidth][c]{
        \includegraphics[width=1.1\linewidth]{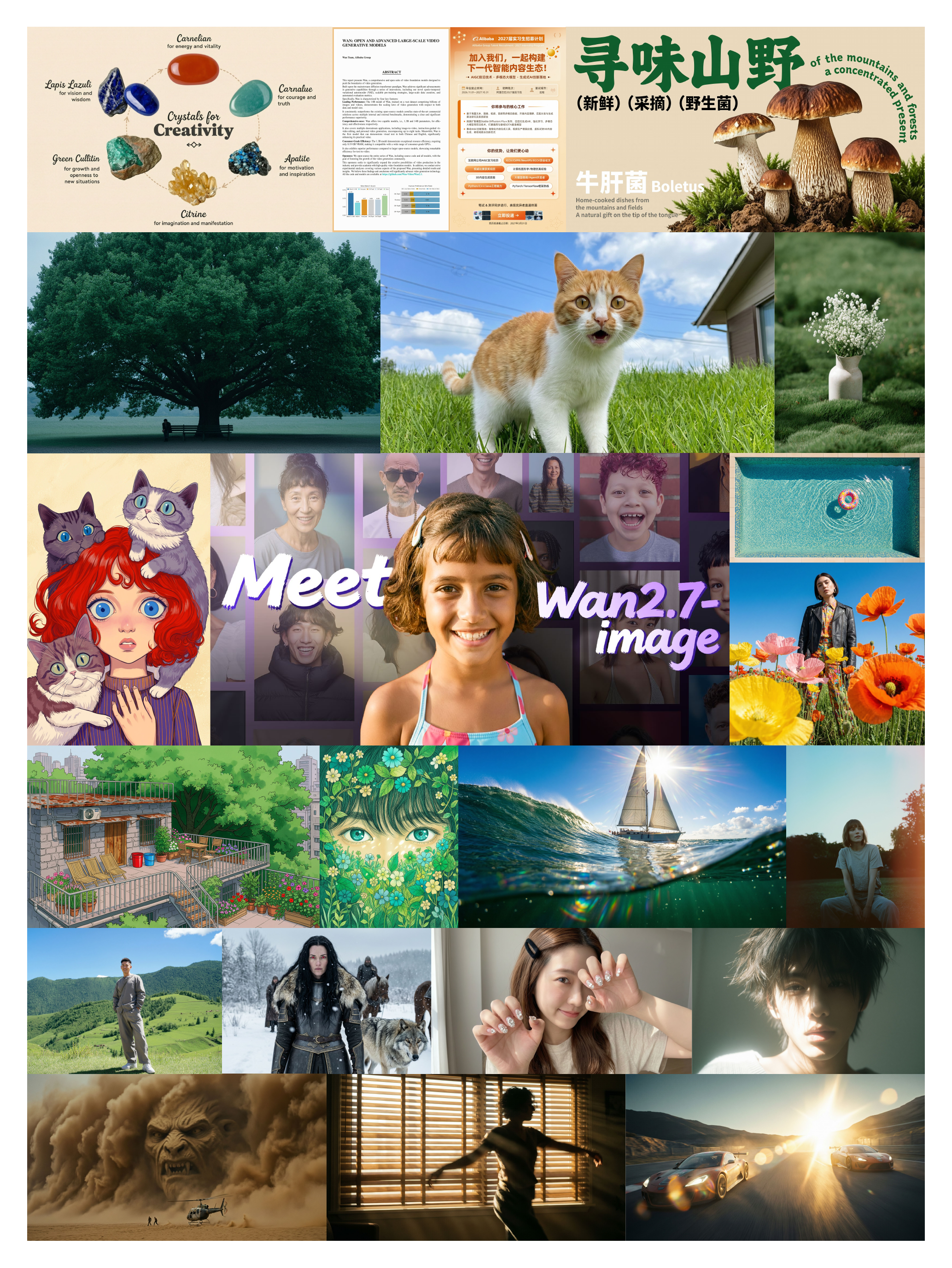}
    }
    \caption{Visual demonstration of diverse capabilities in text-to-image generation.}
    \label{fig:showcase_t2i}
\end{figure}

\clearpage

\begin{figure}[!htb]
    \centering
    \vspace{-20pt}
    \makebox[\linewidth][c]{
        \includegraphics[width=1.1\linewidth]{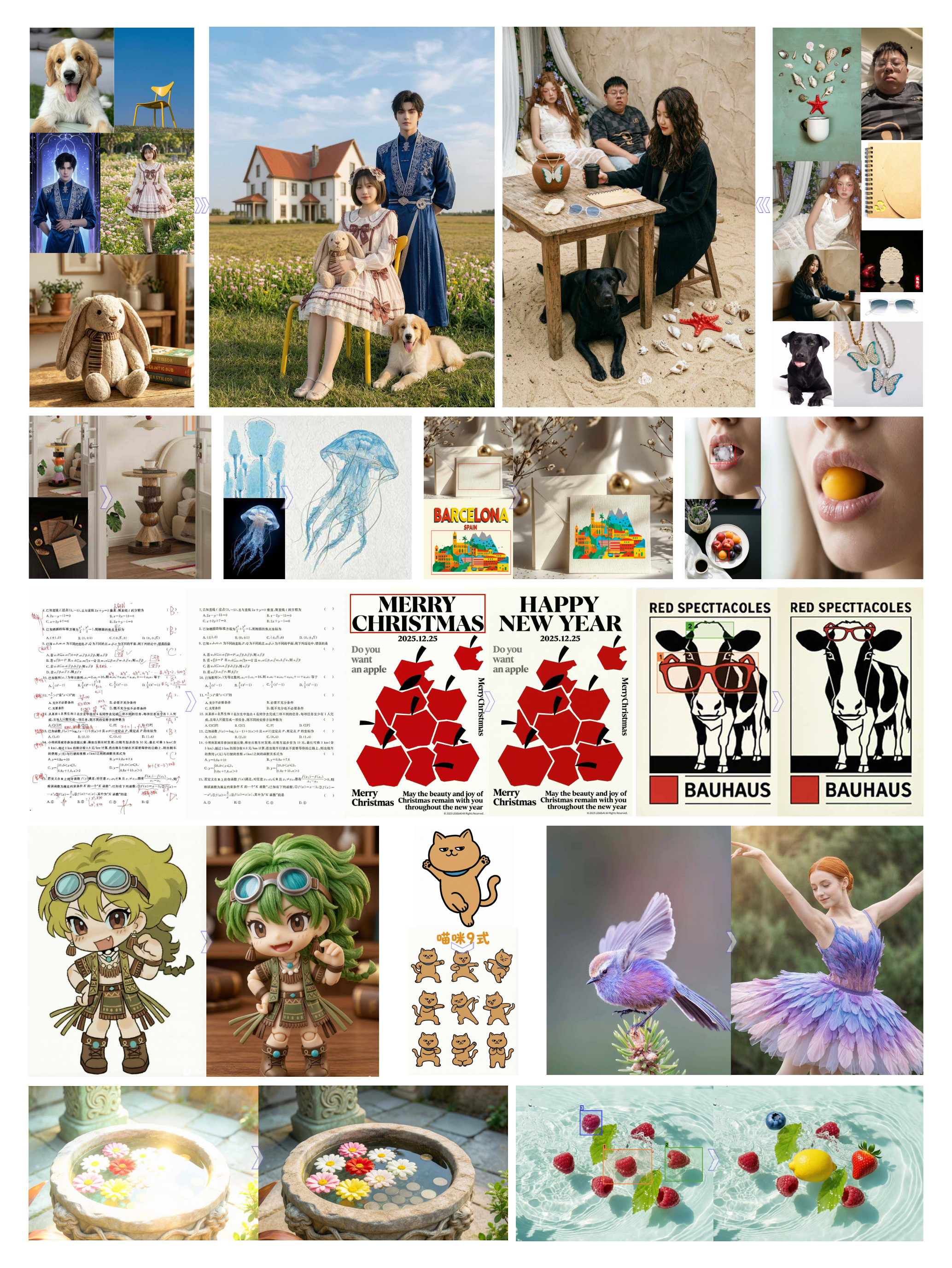}
    }
    \caption{Visual demonstration of diverse capabilities in image-to-image generation (\eg, reference-based generation, instruction-based editing, and interactive editing).}
    \label{fig:showcase_i2i}
\end{figure}

\clearpage

\begin{figure}[!htb]
    \centering
    \vspace{-20pt}
    \makebox[\linewidth][c]{
        \includegraphics[width=1.1\linewidth]{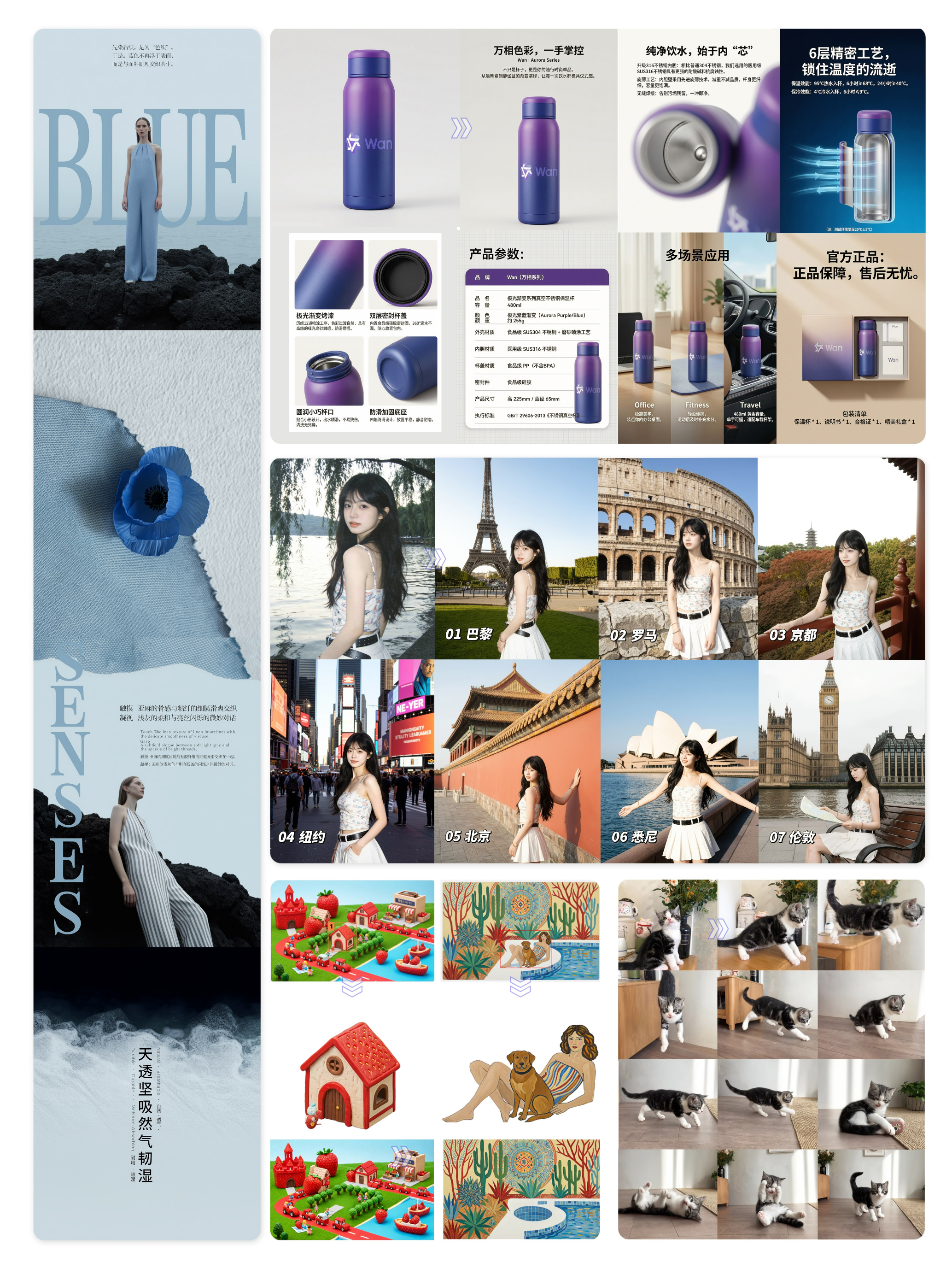}
    }
    \caption{Visual demonstration of diverse capabilities in image series generation.}
    \label{fig:showcase_t2s}
\end{figure}

\clearpage

\begin{figure}[!htb]
    \centering
    \vspace{-20pt}
    \makebox[\linewidth][c]{
        \includegraphics[width=1.1\linewidth]{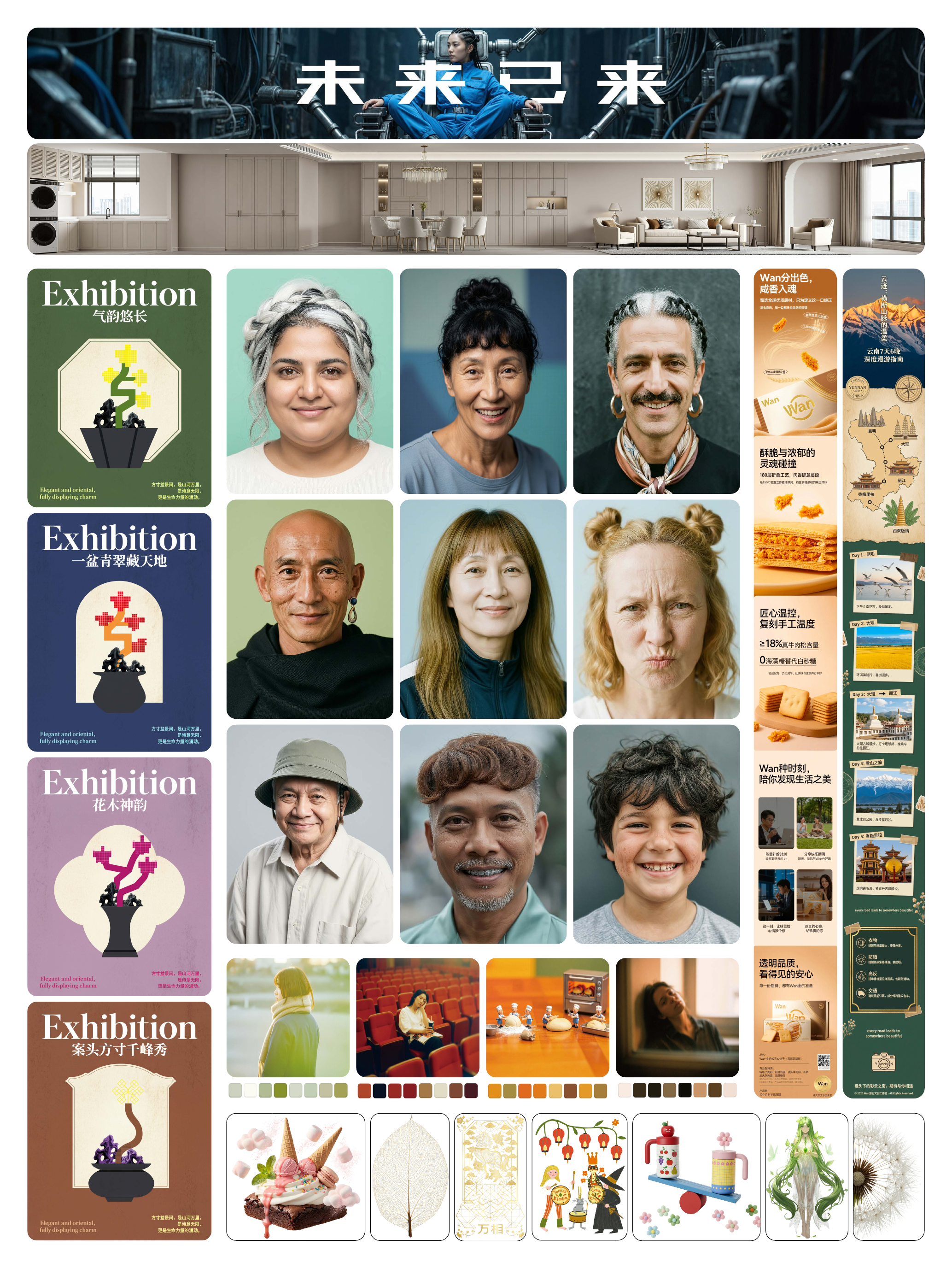}
    }
    \caption{Visual demonstration of diverse capabilities in extreme aspect ratio adaptation, hyper-diverse realistic portrait generation, palette-guided generation, true alpha-channel generation, \etc.}
    \label{fig:showcase_mix}
\end{figure}

\begin{figure}[!htb]
    \centering
    \vspace{-20pt}
    \makebox[\linewidth][c]{
        \includegraphics[width=1.1\linewidth]{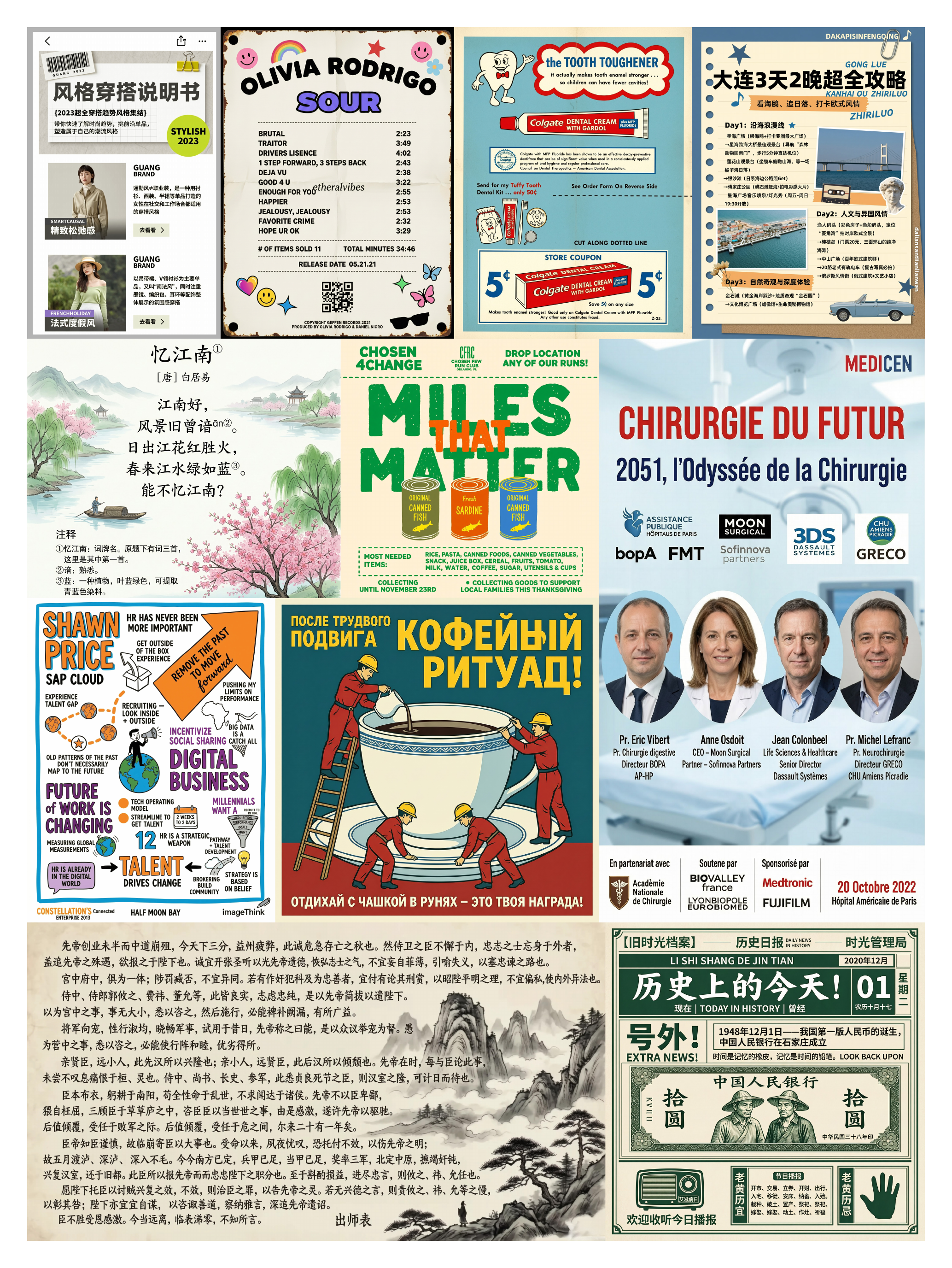}
    }
    \caption{Visual demonstration of diverse capabilities in ultra-long text rendering.}
    \label{fig:showcase_t2i_more_1}
\end{figure}

\clearpage

\section{Introduction}
The past few years have witnessed unprecedented advancements in visual generation~\citep{jiang2025vace, tuo2023anytext, tuo2024anytext2, he2024id,pan2024locate, pan2025ice}, largely driven by the rapid scaling of diffusion models~\citep{peebles2023dit, labs2025flux, esser2024scaling} and the deep integration of multi-modal foundation models~\citep{bai2025qwen25vltechnicalreport, team2023gemini, guo2025seed1p5vl, agarwal2025gpt, yang2025bacon, tang2025unilip, liu2025showtable, liu2025capability}. Contemporary visual generative models such as Seedream series~\citep{seedream2025seedream, gao2025seedream}, HunyuanImage series~\citep{hunyuan2025image2.1, cao2025hunyuanimage3}, and Nano Banana series~\citep{google2026nanobanana2, google2025nanobananapro}, have demonstrated remarkable capabilities in synthesizing high-fidelity and aesthetically stunning images from natural language descriptions.
However, despite their impressive visual quality and enhanced prompt adherence~\citep{han2025ace, mao2025ace++, han2025stylebooth}, most of these cutting-edge systems remain constrained to the realm of casual creation, conceptual exploration, or general-purpose entertainment. When integrated into rigorous, real-world design workflows, even these latest models frequently encounter critical bottlenecks. 
Professional content creation demands significantly more than just aesthetic appeal, and it requires absolute controllability, precise spatial and logical alignment, and the robust execution of complex constraints. Current state-of-the-art models still struggle with tasks that are non-negotiable for professional pipelines, such as rendering ultra-long typography, maintaining strict identity consistency, and executing precise localized interactive editing. Consequently, there is an urgent industry need to paradigm-shift generation models from being mere `image generators' into comprehensive, professional-grade image productivity tools.


To bridge this gap, we introduce \wan, a unified visual generation system explicitly engineered to serve as a next-generation professional productivity tool. Rather than treating image generation as an isolated, open-ended task, \wan is designed to deeply understand and execute complex creative workflows. At its core, the system adopts a unified architecture~\citep{deng2025emerging,liang2024mixture} that conceptually integrates a Planner, an advanced Multi-modal Large Language Model (MLLM) responsible for deep semantic reasoning and task routing, with a Visualizer, a DiT-based module designed for high-fidelity pixel generation~\citep{xing2026wan}, and utilizes a four-channel variational autoencoder (VAE)~\citep{kingma2013vae} to support the generation of transparency channels.
To further eliminate the ambiguity between user inputs and precise visual outputs, we introduce a Prompt Enhancer (PE) that significantly boosts logical reasoning to accurately capture nuanced user intent, alongside an Image Refiner that enhances high-frequency details and pushes the generation resolution to stunning 4K quality. By unifying perception, reasoning, and generation, \wan acts not just as a rendering engine, but as an intelligent design assistant.

A foundational pillar of \wan lies in our unprecedented data scaling and systematic annotation engine. We construct a massive, multi-modal dataset spanning a wide spectrum of tasks, including visual understanding, text-to-image, image-to-image, image series generation, and interleaved generation. To achieve superior cross-modal alignment, we developed a systematic, hierarchical taxonomy and a multi-dimensional fine-grained annotation pipeline. Every image is rigorously categorized and annotated across diverse dimensions. This high-quality data engine is coupled with an advanced, multi-stage training strategy and multi-task optimization. Furthermore, we incorporate meticulously curated reinforcement learning data to optimize the model's outputs for human aesthetic preferences, identity preservation, and complex instruction following.

Empowered by this synergistic design of architecture and data, \wan transcends basic aesthetic generation and instruction following, unlocking a suite of professional productivity capabilities:
\begin{itemize}
    \item \textbf{Ultra-long text rendering:} The ability to accurately render extensive, multi-line typography with complex layouts, rivaling professional typesetting.

    \item \textbf{Extreme aspect ratio adaptation:} Support for generating images with aspect ratios as extreme as 1:8, enabling the creation of expansive cinematic panoramas and ultra-long vertical scrolls.

    \item \textbf{Hyper-diverse realistic portrait generation:} Unprecedented fine-grained facial attribute steering, allowing for highly customizable and consistent virtual character creation.

    \item \textbf{Palette-guided generation:} Fine-grained chromatic control via user-specified hex-code palettes with explicit color proportions.
    
    \item \textbf{Multi-subject identity-preserving generation:} The capability to seamlessly harmonize multiple reference subjects within a single coherent composition.

    \item \textbf{Logical image series generation:} Support for generating a cohesive set of up to 12 images, ensuring rigorous visual consistency and a logical thematic flow across the entire group of samples.
    
    \item \textbf{Multi-modal interactive editing:} Highly precise interactive editing with visual cues that strictly modify targeted regions or extract content from designated reference areas while preserving global context.

    \item \textbf{True alpha-channel generation:} Powered by our novel 4-channel VAE, the model natively generates high-fidelity images with transparent backgrounds, ready for immediate professional design use.

    \item \textbf{High-efficiency 4K generation:} Achieving 4K visual quality without compromising on generation speed.

\end{itemize}

Extensive evaluations demonstrate that \wan achieves highly competitive performance in multi-modal visual generation. As illustrated in Figure~\ref{fig:radar_report}, it comprehensively surpasses Seedream 5.0 Lite and GPT Image 1.5. Although trailing Nano Banana Pro, \wan demonstrates comparable strength in challenging text-to-image scenarios, such as text rendering and photo realism. Furthermore, \wan achieves a pass rate of around 80\% in interactive editing and image series generation, effectively fulfilling real-world productivity requirements.
More generated results of \wan are presented in Figure~\ref{fig:showcase_t2i}, \ref{fig:showcase_i2i}, \ref{fig:showcase_t2s}, \ref{fig:showcase_mix}, and~\ref{fig:showcase_t2i_more_1}.

\section{Data}

\begin{figure}[b]
    \centering
    \includegraphics[width=1.0\linewidth]{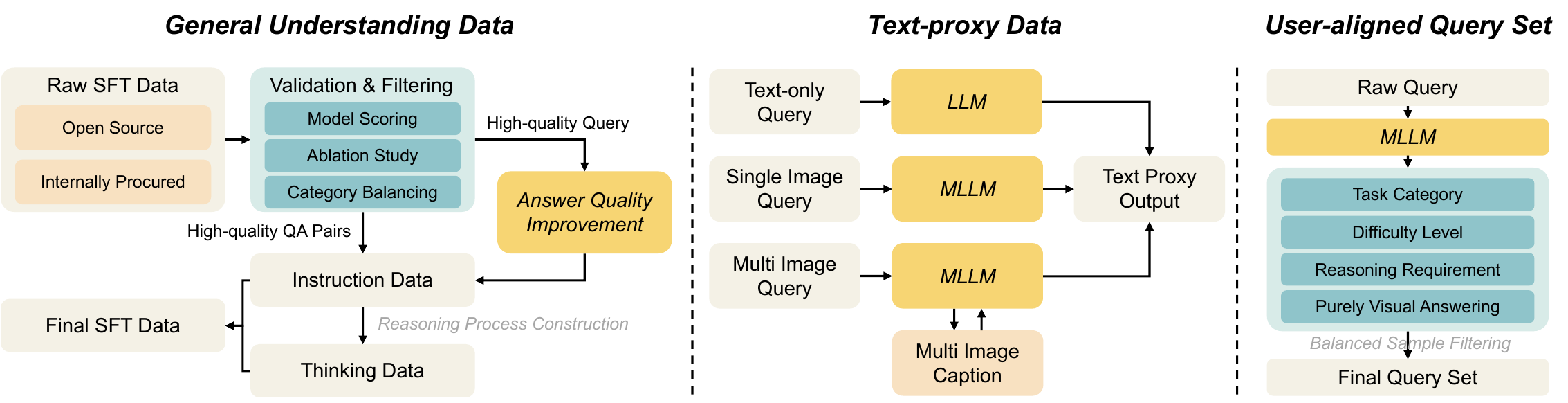}
    \caption{Overview of the construction workflow for understanding data.}
    \label{fig:understanding_data}
\end{figure}

\subsection{Understanding Data}
As depicted in Figure~\ref{fig:understanding_data}, to enhance the understanding capabilities of our model while better supporting a unified multi-modal understanding and generation paradigm, we construct the following understanding-oriented training data~\citep{dan2024topofr, dan2023transface, dan2025transface++}.

\textbf{General understanding data.} To enhance multi-modal understanding capabilities of our model, we construct high-quality data for both text-only and image-text settings. Specifically, we perform quality filtering and category balancing, retain high-quality question-answer pairs, and further improve samples with high-quality queries but low-quality answers by using LLMs and MLLMs to regenerate and filter the responses, yielding a final high-quality instruction-tuning dataset. Furthermore, using the queries from the instruction-tuning data, we employ high-quality LLMs and MLLMs to generate both reasoning processes and final answers. The reasoning processes are enclosed within \texttt{<think>} and \texttt{</think>}. The generated final answers are then compared against the ground-truth answers in the instruction-tuning data, and only reasoning processes that lead to correct answers are retained. This results in a think-augmented training dataset with explicit reasoning supervision.

\textbf{Text-proxy data.} To equip our model with planning capabilities for interleaved generation, we construct text-proxy data with detailed visual prompts. In such text-proxy data, each image placeholder to be generated is marked with \texttt{<BOI>} and accompanied by a detailed description of the expected visual content. These visual prompts are wrapped with \texttt{<imagine>} and \texttt{</imagine>}, providing richer image-specific guidance compared to the surrounding sparse textual context. Specifically, the data fall into three categories. The first consists of 

text-only inputs (\eg, category labels or keywords), where we use the text input as the user query and employ a high-quality LLM to construct the text proxy. The second consists of single-image inputs, where we feed the image directly into a high-quality MLLM and prompt it to jointly construct both the user query and the text proxy. The third consists of multiple correlated images, where we first generate per-image descriptions using an MLLM, then feed both the images and descriptions back into the MLLM to organize them into a coherent interleaved narrative, followed by a refinement step for logical and stylistic consistency.

\textbf{User-aligned query set.} This subset is designed to improve our model's real-world usability. We employ a high-quality MLLM to annotate each raw query across a set of attributes, including task category, difficulty level, whether numerical or logical reasoning is required, and whether the query is purely a visual question answering. Based on these annotations, we apply manually defined sampling rules to construct a final query set that is as balanced as possible across all attribute dimensions.

\subsection{Generation Data}
\begin{figure}
    \centering
    \includegraphics[width=1\linewidth]{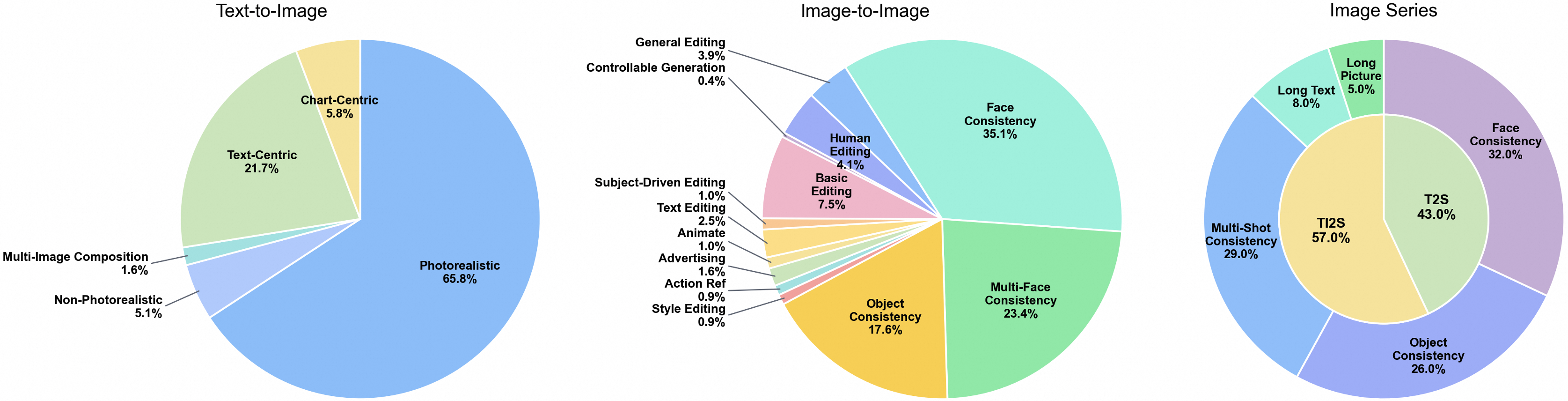}
    \caption{Dataset category distribution (\%) for Text-to-Image, Image-to-Image, and Image Series.}
    \label{fig:data_distribution}
\end{figure}

Our training data systematically covers the full spectrum of generative tasks. In Text-to-Image (T2I), our model excels in accurate long-text rendering and supports extreme aspect ratios of up to 1:8. For Image-to-Image (I2I), we incorporate a diverse range of editing and reference-based tasks, accommodating up to 9 input images for complex conditioning. Furthermore, in Text-to-Image-Series (T2S) and Text-Image-to-Image-Series (TI2S), our framework enables the synthesis of up to 12 images, prioritizing rigorous visual consistency and logical thematic progression across the entire series. The overview of the dataset distribution is shown in Figure~\ref{fig:data_distribution}.

\subsubsection{Data Collection}
To filter high-quality training datasets from large-scale data, we design and deploy an efficient data retrieval system together with fine-grained, multi-dimensional image operators that significantly improve both the efficiency of data selection and the overall quality of the resulting dataset. The workflow and its core components are outlined as follows.

\begin{figure}[!ht]
    \centering
    \includegraphics[width=0.75\linewidth]{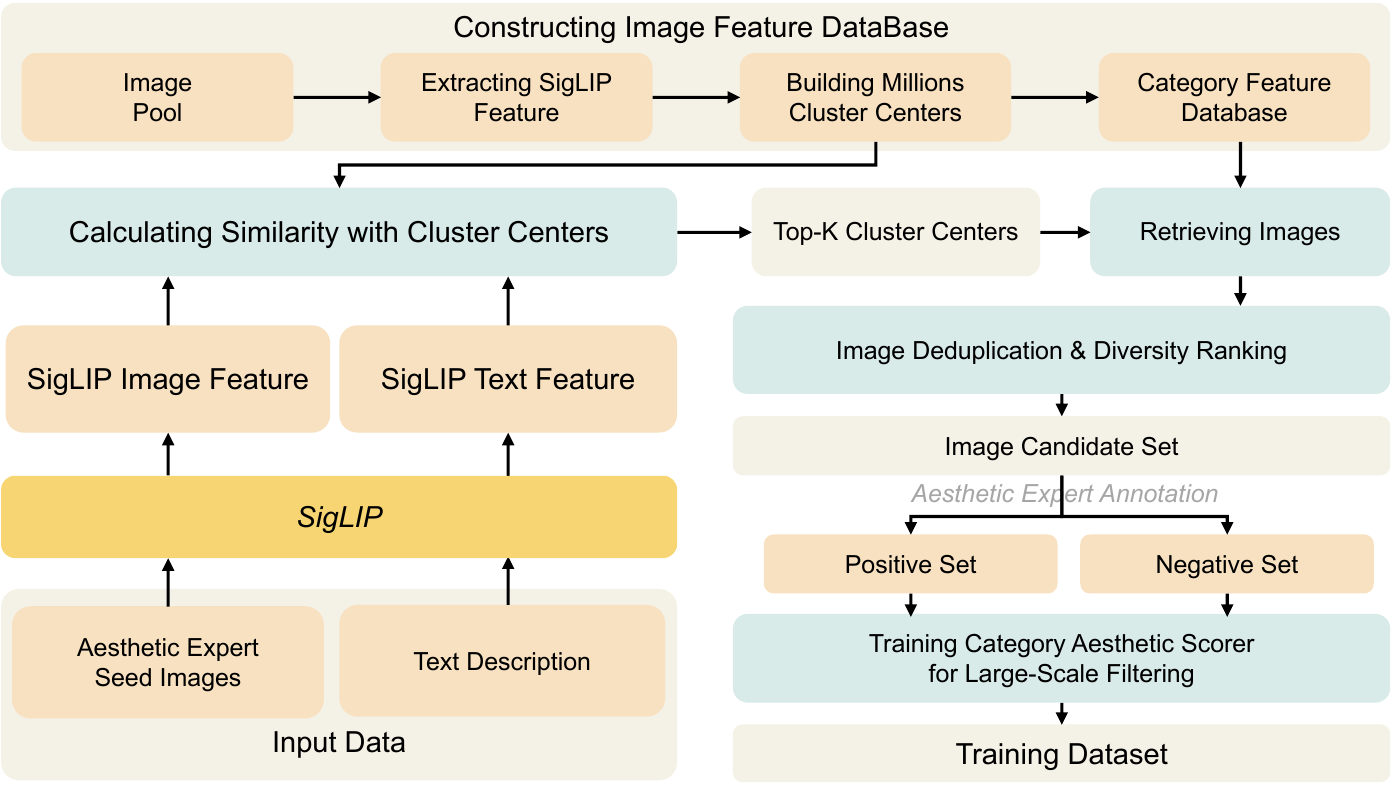}
    \caption{Overview of the multi-modal retrieval system.}
    \label{fig:data_retrieval}
\end{figure}
\paragraph{Multi-modal retrieval system.} As shown in Figure~\ref{fig:data_retrieval}, to improve the efficiency of acquiring high-quality data, we develop a multi-modal retrieval system, which establishes a closed-loop pipeline integrating data retrieval, annotation, and cleaning.
Training the model requires large volumes of high-quality data that are strongly associated with specific target categories. To address the challenges of low acquisition efficiency and limited relevance in collecting data for key categories, we develop a multi-modal retrieval system, forming a positive feedback loop of image retrieval, annotation, iterative retrieval, and data cleaning.
During collecting key-category data, the defining characteristics of many categories often lie in aesthetic attributes or compositional logic, which are difficult to express and reliably retrieve using conventional single-image search. To address this limitation, the system introduces multi-image retrieval capabilities, where multiple seed images jointly define the target distribution, thereby significantly improving both the relevance and controllability of the retrieved results.
The system supports multiple retrieval modes over large-scale image repositories, including image-to-image search, multi-image search, text-to-image search, image–text hybrid search, and batch retrieval. In addition, a cluster-based diversity re-ranking strategy is employed to improve the coverage and diversity of retrieved results while maintaining strong relevance.

\paragraph{Multi-dimensional operators.} To comprehensively quantify image quality, we construct a set of fine-grained, multi-dimensional image operators spanning five mutually independent yet complementary dimensions: image feature extraction, aesthetic quality assessment, AI-generated content detection, low-level information evaluation, and overall image quality assessment.

Specifically, we begin with a fine-grained image feature extraction operator, which provides robust and reliable representations for subsequent quality evaluation tasks. Building on these representations, we introduce a multi-level aesthetic quality assessment operator to quantify aspects such as composition, color distribution, and overall visual appeal, thereby facilitating the selection of images that are clearer, more natural, and better aligned with human aesthetic preferences. In parallel, we incorporate a multi-level AI-generated content detection operator to efficiently identify and filter images with more realistic visual characteristics and a lower likelihood of being AI-generated. To further improve data quality from the perspective of low-level visual information, we introduce a low-level information evaluation operator based on information entropy and compression artifact detection, which helps remove low-quality samples characterized by low information density, limited content diversity, insufficient effective detail, or severe compression distortion. Beyond these criteria, we further incorporate an overall image quality assessment operator that leverages global indicators such as watermark detection and greasy-texture detection, enabling more comprehensive quality assessment and cleaning of large-scale image datasets.

\begin{figure}[t]
    \centering
    \includegraphics[width=0.95\linewidth]{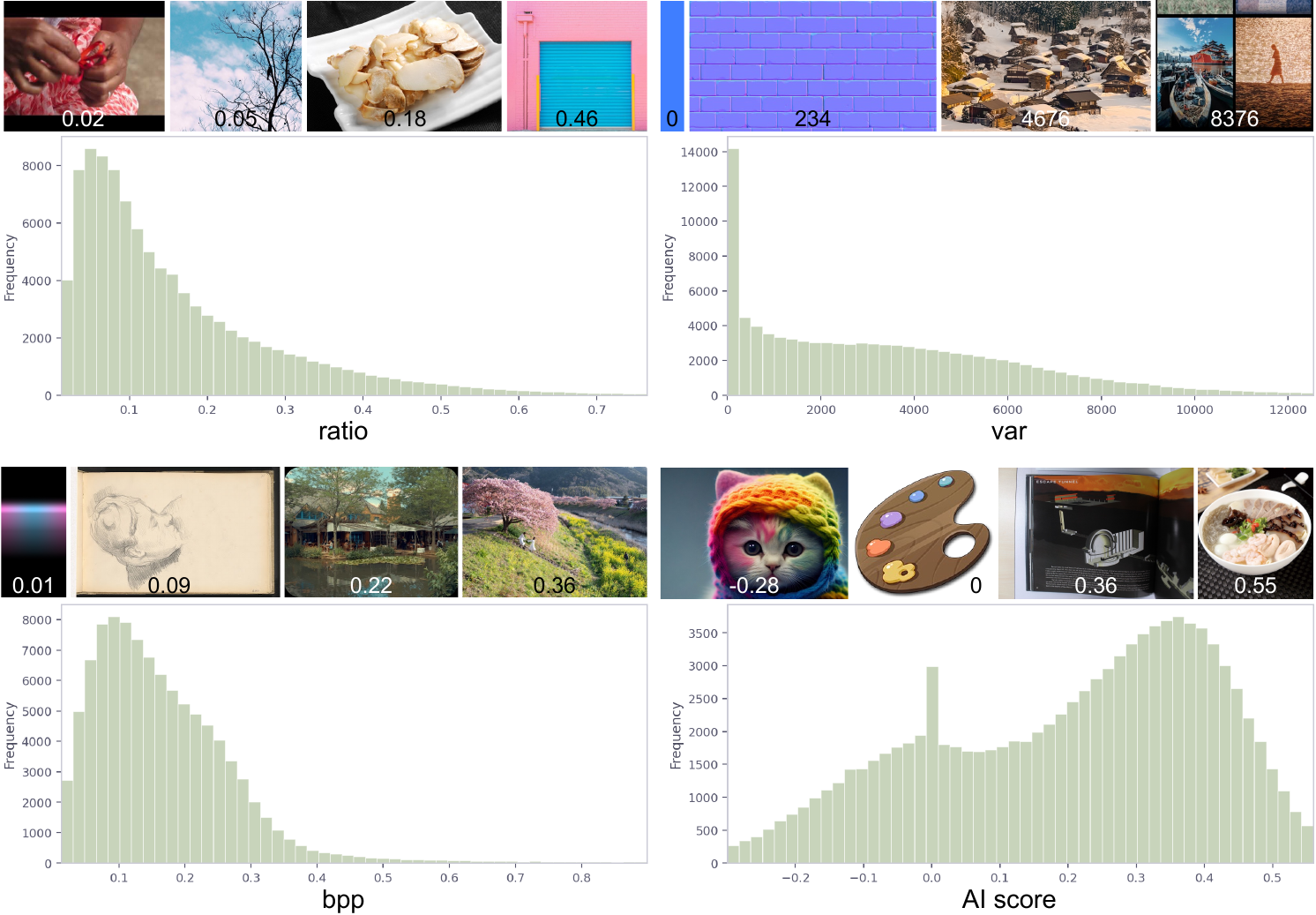}
    \caption{Distribution and visual examples of the representative filtering operators.}
    \label{fig:filtering_operator}
\end{figure}

Through these coordinated multi-dimensional evaluation operators, we significantly improve the efficiency of data filtering while enhancing the clarity, realism, aesthetic quality, and overall usability of the resulting dataset. Figure~\ref{fig:filtering_operator} visualizes the distributions and examples of some representative operators:
\begin{itemize}
 \item \textbf{Compression artifact (ratio)}: To identify excessively compressed images, we compute the ratio between the theoretical uncompressed file size (derived from image resolution and bit depth) and the actual file size. A lower ratio indicates a higher degree of compression, which serves as a proxy for potential quality degradation and loss of fine-grained details.
\item \textbf{Edge pixel variance (var)}: To maximize the density of informative visual content, we filter out low-entropy images using an entropy-inspired criterion. Specifically, we measure the variance of edge pixels to identify images with large homogeneous backgrounds or prominent solid-color borders, as such regions contribute little to effective feature learning.
\item \textbf{Image information complexity (bpp)}: 
To identify and exclude images with low information density, we use a JPEG-based Complexity Proxy. Specifically, we re-encode each image into JPEG format and compute the corresponding bits-per-pixel, which serves as an approximation of structural complexity.
Images with low bpp are filtered out, as they generally contain limited textural information.
 \item \textbf{Artificiality score (AI score)}: We employ a specialized AI-detection classifier to evaluate the authenticity of each image. The AI score reflects the probability of an image being authentic (i.e., not AI-generated), a higher score denotes greater authenticity, while a lower score indicates a higher likelihood that the image is synthetically generated.
\end{itemize}

\subsubsection{Structured Data Taxonomy and Captioning}
\begin{figure}[ht]
    \centering
    \includegraphics[width=0.8\linewidth]{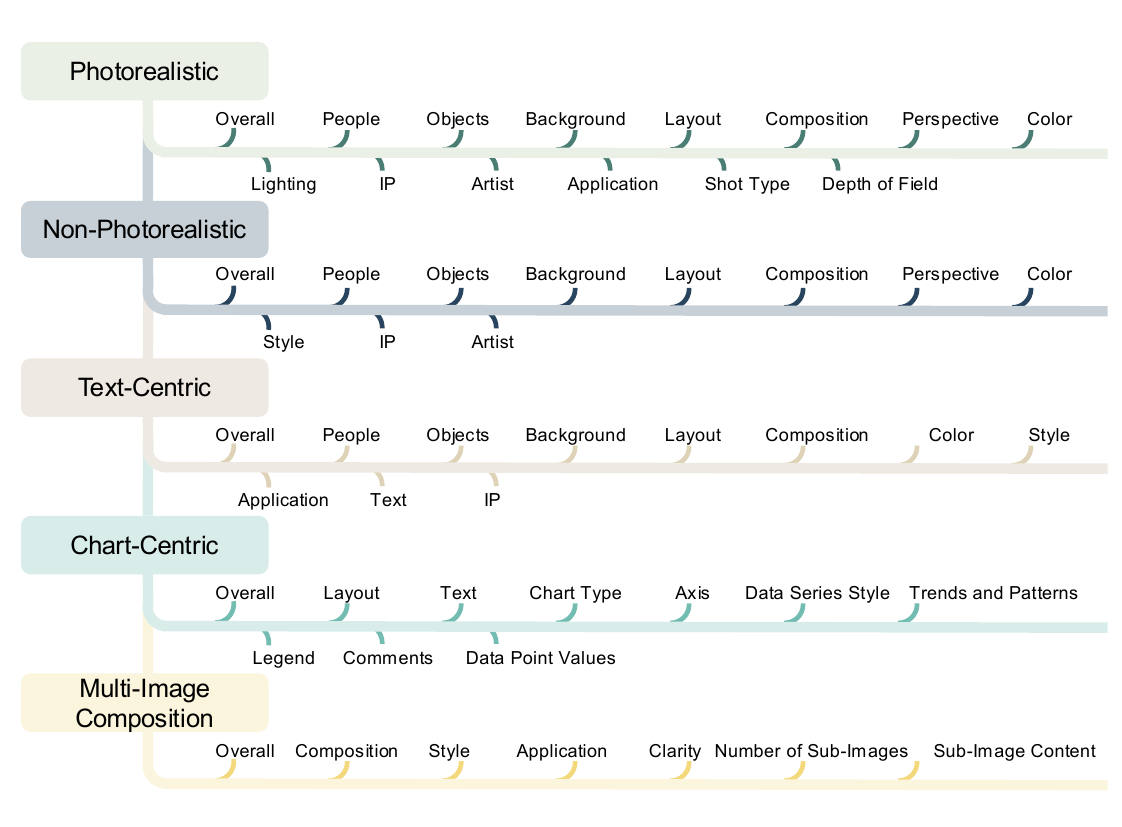}
    \caption{Overview of the structured data taxonomy. Our framework routes images into five primary categories and assigns them category-specific attributes (\eg, Lighting for photorealistic images, Axis for charts).}
    \label{fig:data_taxonomy}
\end{figure}

\paragraph{Structured data taxonomy. }
To ensure a balanced distribution in modeling general generative capabilities, we first introduce a structured data taxonomy for training images, organizing and controlling sampling across dimensions such as semantic themes, image attributes, style categories, compositional patterns, and generation difficulty, as shown in Figure~\ref{fig:data_taxonomy}. Such a design improves data coverage and mitigates common distribution biases. For image-conditioned tasks, such as {I2I} and {TI2S}, we further construct a fine-grained task labeling system. Specifically, we establish a hierarchical taxonomy over dimensions including the reference or editing paradigm, editing target, and editing type, thereby enabling more effective organization of complex editing tasks and alleviating the training imbalance caused by long-tail samples.

\paragraph{Structured data captioning.}
We train the model on diverse annotations for robust instruction following across text prompts of varying descriptive granularities. Our captions comprise three main categories:
\begin{itemize}
    \item \textbf{Raw captions.} The raw caption comprises the original, uncurated textual metadata harvested alongside the image from its source environment. This includes information such as user-defined tags, HTML alt-texts, webpage titles, and contextual snippets from the surrounding anchors or articles. Unlike human-annotated descriptions, raw captions preserve the naturally occurring linguistic noise and weak supervision signals inherent in web-scale data.

    \item \textbf{Natural language captions.} We design a variety of prompting templates and employ advanced MLLMs to generate image descriptions with multiple granularities, ranging from concise captions to detailed, multi-sentence paragraphs.

    \item \textbf{Structured JSON captions.} We design category-specific structured captions to provide multi-dimensional and fine-grained image descriptions, thereby endowing the model with highly precise instruction understanding and detail reconstruction capability. The structured captions possess the following key characteristics.
\textbf{(i)} Image classification and routing mechanism: Prior to fine-grained annotation, each image is strictly categorized into one of five fundamental classes: {photorealistic images}, {non-photorealistic images}, {text-centric images}, {charts}, and multi-image compositions.
\textbf{(ii)} Dynamic dimension mapping: Each fundamental category is associated with a distinct set of fine-grained annotation dimensions. The global annotation pool comprises 25 independent attributes, including global semantics, human subjects, objects and props, background environment, spatial layout, visual composition, and others.
\textbf{(iii)} Timely update mechanism: Structured captions are updated in a timely manner to incorporate missing dimensional information and remove inaccurate descriptions from specific dimensions, while preserving all other valid dimension-wise information, thereby ensuring the accuracy and completeness of the descriptions.
\textbf{(iv)} Natural language rewriting: An MLLM is fine-tuned to integrate the dimension-wise descriptions provided in the structured captions and generate natural language rewrites that capture rich fine-grained details.
\end{itemize}

\subsubsection{Stage-wise Training Data Strategy}
During the whole model training stage, we moderately upweight the sampling ratio of tasks that typically require longer training, such as text generation and identity preservation. Meanwhile, we construct multiple sets of detailed prompts to enhance the alignment between textual semantics and visual representations. 
As the training trajectory progressively evolves from the initial Pre-training (PT) and Continual Pre-training (CT) stages to the final Supervised Fine-tuning (SFT), we strategically and dynamically rebalance sampling ratios across diverse task categories while incorporating carefully curated high-quality data to bolster and refine specialized capabilities. In the concluding SFT stage, we retain only exemplary data that passes rigorous quality and aesthetic filtering to further elevate the visual fidelity of model outputs. By leveraging this systematic multi-stage data scheduling, the model simultaneously attains precise semantic instruction following and superior aesthetic performance.

\paragraph{Cross-stage data construction and diversity.}
For the PT stage, we utilize a multi-modal retrieval system to curate massive-scale datasets, ensuring broad semantic coverage. 
For the CT and SFT stages, we employ MLLMs for high-precision tagging to optimize key capabilities, including stylistic diversity, conceptual rendering, scene text, and visual aesthetics, ensuring the model possesses profound semantic understanding. 
Specifically, we meticulously categorize the annotated content and address data balancing for long-tail categories, leveraging fine-grained captioning to further enhance the alignment between complex textual prompts and visual representations.

\paragraph{Multi-dimensional filtering and stage-wise scheduling.}
Across all training stages, we implement a rigorous filtering pipeline to ensure data purity and aesthetic excellence:
\textbf{(i)} Feature extraction and deduplication: We extract parallel features, including SigLIP-2~\citep{tschannen2025siglip2} semantic embeddings and low-level metadata, to facilitate feature-level deduplication of same-source images and maximize dataset diversity.
\textbf{(ii)} Comprehensive quality assessment: Each sample is quantitatively scored across over 20 dimensions, such as aesthetic quality, AI-generated artifact detection, over-smoothing assessment (\eg, greasy textures), and watermark detection.
\textbf{(iii)} Stage-wise dynamic thresholding: Filtering thresholds are adaptively adjusted based on the training stage; while the PT stage maintains a focus on scale with baseline quality filters, the CT and SFT stages implement significantly more stringent thresholds for aesthetics and clarity to ensure high-fidelity artistic outputs.
\textbf{(iv)} Manual verification: Post-automated filtering, random samples are manually inspected to verify the robustness of the scoring framework and ensure that only the highest-quality data contributes to the final training stages.

\section{Model}
\begin{figure}
    \centering
    \includegraphics[width=1.0\linewidth]{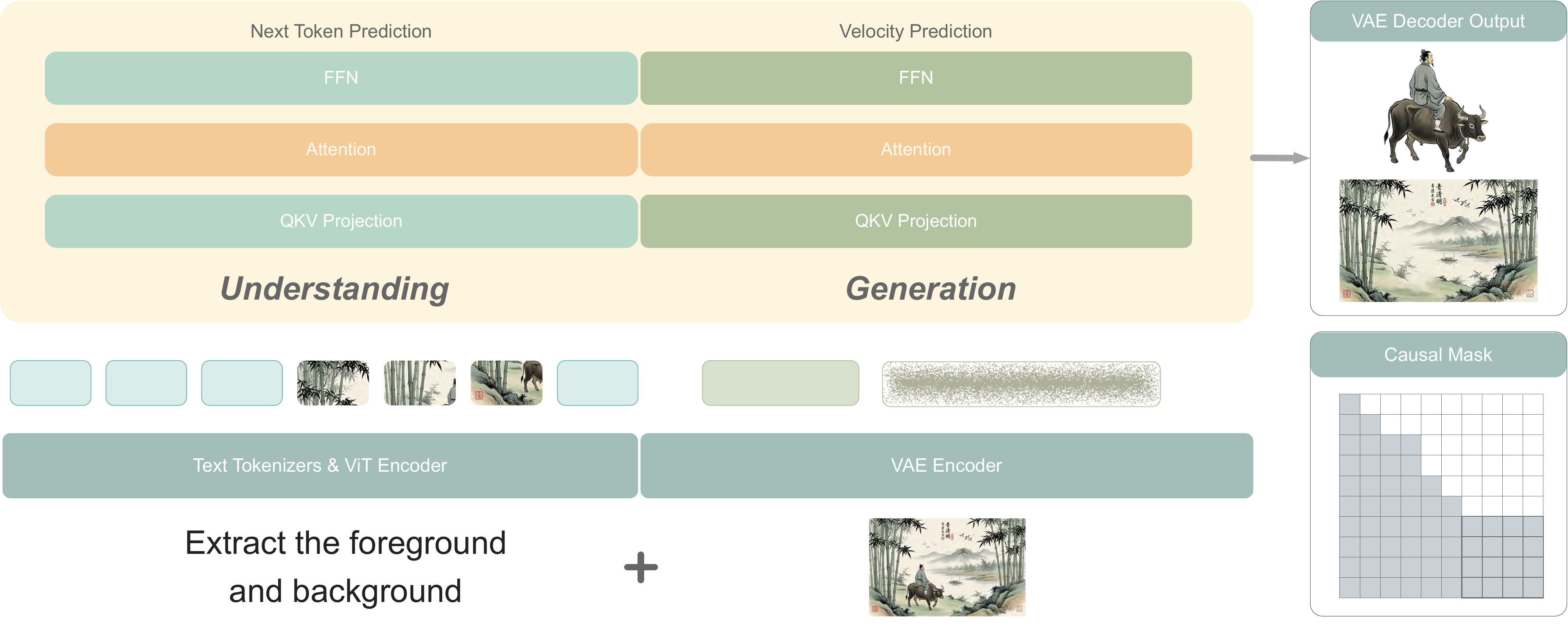}
    \caption{
    Overview of the \wan architecture.
    }
    \label{fig:framework}
\end{figure}

\subsection{Architectural Design}
As shown in Figure~\ref{fig:framework}, this model adopts a unified Transformer architecture for unified multi-modal understanding and generation, where dedicated Transformer experts share attention within each block for efficient feature space alignment. The understanding branch adopts the standard Decoder-only Transformer architecture, while the generation branch adopts a DiT~\citep{peebles2023dit} architecture trained under the Rectified Flow~\citep{liu2023flow} paradigm, with LayerNorm~\citep{ba2016layernorm}, SiLU~\citep{elfwing2018silu}, and QK-Norm~\citep{henry2020qknorm} incorporated to enhance training stability. We adopt a decoupled training strategy that optimizes the two branches in separate stages, with the understanding branch trained first. The attention weights of the generation branch are then initialized from the well-trained understanding branch to ensure representational consistency, while newly introduced modules are randomly initialized. This strategy enables independent optimization of each branch while preserving unified inference under the unified Transformer architecture.

\paragraph{Attention and modulation mechanism.}
Within the unified architecture, we introduce a generalized attention mechanism with a dedicated modulation strategy. To preserve auto-regressive generation, we adopt a structured attention mask: causal attention~\citep{yang2021causal} is applied among text, ViT~\citep{dosovitskiy2020vit}, and VAE tokens, while bidirectional attention~\citep{wibisono2023bidirectional} is used within the visual tokens of each image. For image series generation, bidirectional attention is further enabled across the visual tokens of all images in the same series, promoting identity consistency and style coherence across images. Furthermore, to preserve fine-grained details in image editing tasks, the clean VAE tokens encoded from the input image are fed into the DiT as additional conditional inputs. Both the clean and noised VAE tokens receive timestep embeddings and are used to modulate each DiT block via AdaLN~\citep{huang2017arbitrary}. Inspired by Teacher Forcing~\citep{cho2014teacherforcing}, the timestep of the clean VAE tokens is explicitly set to zero, treating them as deterministic conditions to retain the structural and content characteristics of the input image.

\paragraph{Positional encoding.}
In the understanding branch, the ViT module serves as the visual encoder, converting image pixels into visual tokens, which are then positionally encoded alongside text tokens via Multi-modal Rotary Positional Embedding (MRoPE)~\citep{huang2025mrope}. In the generation branch, we instead employ a 3D-RoPE scheme, assigning different positional sensitivities to the temporal dimension $T$ and spatial dimensions $H$ and $W$, with a fixed positional offset inserted along the $T$ axis between text and image segments as a semantic boundary. To bridge the two branches, contextual representations from the understanding module are seamlessly integrated with those of the generation branch, establishing a unified reference frame for positional alignment. 3D-RoPE is then applied to both the queries of the generation tokens and the integrated representations, ensuring consistent relative positional encoding across the two branches. This enables the visual tokens in the generation branch to attend over a shared spatio-temporal context, achieving consistent positional alignment between understanding and generation tokens.

\subsection{VAE}

In visual generative models, the quality of the VAE mapping between image space and latent space largely determines the upper bound of generation fidelity. Although existing VAEs perform well on natural images, they still struggle in scenarios that require precise high-frequency reconstruction, alpha-channel modeling, and super-resolution. These limitations are particularly evident in small-font text, dense layouts, complex textures, fine boundaries, and transparent regions, where artifacts such as blurred characters, merged strokes, boundary contamination, and distorted alpha transitions frequently occur. To address these issues, we propose a high-fidelity four-channel VAE that jointly models RGB content and alpha transparency, providing higher-quality latent representations for image generation and editing.

\paragraph{Architecture design and training loss.}
In the architecture design, we explicitly balance compression efficiency against reconstruction fidelity. At the input stage, we adopt a $2\times2$ patchify scheme, followed by an $8\times8$ spatial downsampling pipeline in the backbone, resulting in an overall compression ratio of $16\times16$. Building on this design, we stack multiple residual blocks across multiple stages of both the encoder and decoder to strengthen deep nonlinear representation capacity. This architecture preserves computational efficiency while enabling high-fidelity reconstruction of RGBA images, showing particularly stable recovery of transparent boundaries, semi-transparent regions, and fine-grained textures.

Compared with conventional VAEs designed primarily for RGB images, our method explicitly models the alpha channel, making it better suited for transparent image content such as 4-channel PNGs. Such data are widely used in practical applications, including e-commerce assets, game resources, and layered design elements. The main challenge lies not only in reconstructing foreground appearance, but also in accurately preserving transparent regions, semi-transparent transitions, and boundary compositing effects. RGB-only modeling often fails to disentangle background color from transparency, leading to artifacts such as black or white fringes, aliasing, halos, and background leakage around transparent boundaries. To address this issue, we introduce a hybrid reconstruction loss that jointly constrains visible-content reconstruction, alpha structure recovery, and transparent-boundary quality, facilitating more natural opacity transitions and cleaner object contours.

\paragraph{Training strategy.}
We adopt a three-stage progressive training schedule to enable stable adaptation from low to super resolutions. In the first stage, the model is trained on $256\times256$ images, which strengthens its perception of transparent content and basic structure. In the second stage, we switch to mixed-resolution training, improving generalization across multi-scale visual patterns. In the third stage, we further introduce a 4-channel GAN discriminator on top of mixed-resolution training, in order to improve detail sharpness, transparent-boundary realism, and overall visual consistency. Through this curriculum, the model gradually adapts from 256$\times$256 to 2K resolution and maintains strong stability and generalization in high-resolution RGBA modeling.

In addition, following VA-VAE~\citep{yao2025vavae}, we incorporate semantic distillation during training to enrich the latent representation with high-level semantic information. With the CT stage on large-scale and diverse data, the model develops a stronger cross-domain detail modeling ability. The training data cover plenty of challenging visual content, including multilingual scene text, document text, synthetic typographic text, dense face samples, and complex textured images. This diversity enables the model to more accurately capture fine-grained textual structures, local facial details, and repetitive texture patterns. We empirically observe that, as the model improves its understanding of high-frequency local structures, it gradually acquires tiny text reconstruction capability. In particular, it can recover small characters, thin strokes, and densely arranged layout details with substantially improved clarity, significantly alleviating the blurred text, broken strokes, and character merging commonly observed in conventional VAEs. This capability is quite important for several tasks, such as text generation, document reconstruction, layout generation, and multilingual visual content modeling. Therefore, it provides a strong foundation for improving text readability and layout stability in downstream generative models.

To evaluate the proposed VAE, we compare it against recent state-of-the-art methods on RGB and RGBA images at different resolutions, including FLUX.1 VAE~\citep{labs2025flux}, SD3.5 VAE~\citep{esser2024sd}, Qwen-Image-Layered VAE~\citep{yin2025qwenimagelayered} , HunyuanImage-3.0 VAE~\citep{cao2025hunyuanimage3}, Wan2.2 VAE~\citep{wan2025wan22}, and AlphaVAE~\citep{wang2025alphavae}. The results are shown below.

\begin{table}[t]
\centering
\caption{Reconstruction comparison on 5,000 RGB images at 512p resolution. The best results are in \textbf{bold}.}
\label{tab:rgb_recon}
\setlength{\tabcolsep}{9pt}
\begin{tabular}{lccccc}
\toprule
Method & Spatial Downsampling & Latent Dim. & PSNR$\uparrow$ & SSIM$\uparrow$ & LPIPS$\downarrow$ \\
\midrule
FLUX.1 VAE              & $8\times8$   & 16 & 34.769 & \textbf{0.958} & \textbf{0.009} \\
SD3.5 VAE            & $8\times8$   & 16 & 33.000 & 0.942          & 0.012 \\
Qwen-Image-Layered VAE         & $8\times8$   & 16 & 33.447 & 0.946          & 0.020 \\
HunyuanImage-3.0 VAE & $16\times16$ & 32 & 33.735 & 0.939          & 0.022 \\
Wan2.2 VAE           & $16\times16$ & 48 & 33.708 & 0.946          & 0.014 \\
\rowcolor{Gray}
\textbf{Wan-Image VAE} & $16\times16$ & 48 & \textbf{35.429} & \textbf{0.958} & 0.010 \\
\bottomrule
\end{tabular}
\label{tab:vae_1}
\end{table}
\begin{table}[t]
\centering
\caption{Reconstruction comparison on 500 RGBA images at 1080p resolution. The best results are in \textbf{bold}.}
\label{tab:rgba_recon}
\setlength{\tabcolsep}{9pt}
\begin{tabular}{lccccc}
\toprule
Method & Spatial Downsampling & Latent Dim. & PSNR$\uparrow$ & SSIM$\uparrow$ & LPIPS$\downarrow$ \\
\midrule
AlphaVAE (FLUX.1)
& $8\times8$   & 16 & 36.797          & 0.970          & 0.035 \\
Qwen-Image-Layered VAE     & $8\times8$   & 16 & 37.884          & \textbf{0.971} & \textbf{0.018} \\
\rowcolor{Gray}
\textbf{Wan-Image VAE} & $16\times16$ & 48 & \textbf{38.274} & 0.970 & \textbf{0.018} \\
\bottomrule
\end{tabular}
\label{tab:vae_2}
\end{table}

\begin{figure}[!t]
    \centering
    \includegraphics[width=1.0\textwidth]{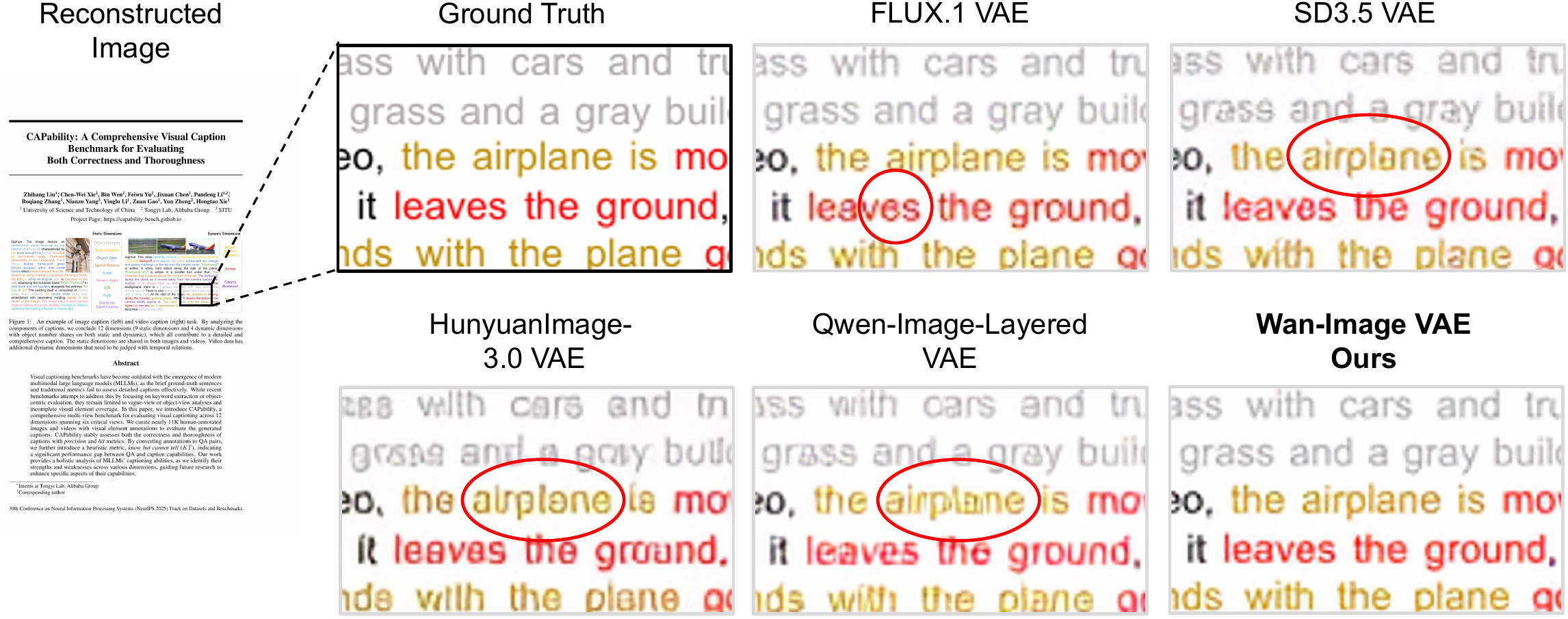}
    \caption{Visual comparison of different VAE models. Wan-Image VAE preserves clearer character-level details than existing methods.}
    \label{fig:vae_visual}
\end{figure}

\paragraph{Quantitative comparison.}
As shown in Table~\ref{tab:vae_1}, we evaluate reconstruction quality on 5,000 RGB images. The proposed VAE achieves the best overall performance, attaining the highest PSNR of 35.429 and SSIM of 0.958 under $16{\times}16$ spatial downsampling, while maintaining a low LPIPS of 0.010. These results demonstrate that even under a high compression ratio, our VAE effectively preserves structural and fine-grained image details, with reconstructions exhibiting strong perceptual similarity to the original images, reflecting superior latent space expressiveness and decoding quality.
In addition, as depicted in Table~\ref{tab:vae_2}, we evaluate reconstruction quality on 500 RGBA images. The proposed VAE achieves the highest PSNR among all compared methods, demonstrating superior pixel-level reconstruction accuracy over existing RGBA VAE approaches and confirming its strong compression and reconstruction capability in RGBA settings.

\paragraph{Qualitative comparison.}
We present qualitative comparisons on document regions with fine-grained text in Figure~\ref{fig:vae_visual}. The results show that our VAE better preserves fine character strokes and glyph boundaries in document regions, yielding reconstructions with higher legibility. The proposed VAE demonstrates superior reconstruction of high-frequency details, recovering sharper and more stable local structures even under higher compression ratios.

\subsection{Planner and Visualizer}
Our model establishes a unified architecture for multi-modal understanding and generation, integrating an MLLM-based semantic understanding stream and a DiT-based visual generation stream within a single model. To empower the model with the capacity for autonomous decision-making and automated planning, we construct a unified pipeline that achieves a seamless transition from semantic understanding to visual generation through the synergetic collaboration between an MLLM-based Planner and a DiT-based Visualizer~\citep{xing2026wan}. This design enables automatic task switching across a broad range of multi-modal tasks for arbitrary user prompts, including language understanding, visual understanding, image generation, image editing, and interleaved multi-modal generation.

Furthermore, we support Think Mode across all tasks to improve user intent understanding. For generation-related tasks, including T2I, I2I, and interleaved multi-modal generation, we design a CoT-driven planning mechanism. The user's generative intent is translated into dense textual descriptions or structured instructions, represented by special tokens \texttt{<imagine>} and \texttt{</imagine>} producing richer outputs grounded in world knowledge.
Simultaneously, we design a mechanism to dynamically infer the output resolution from user input prompts, including aspect ratio, layout orientation, and scene description. The inferred resolution is encoded as special tokens \texttt{<size>grid\_h*grid\_w</size>}. In multi-image input I2I settings, the resolution is further inferred from the provided editing or reference images, enabling optimal alignment with user intent.

For interleaved multi-modal generation, the model autoregressively generates text tokens while producing precisely interleaved images via visual CoT planning. For example, given a prompt such as `generate an illustrated fable about the tortoise and the hare with interleaved images and text', the model automatically identifies the task type and initiates the interleaved generation process. Specifically, the Planner continuously generates text tokens and predicts a \texttt{<BOI>} token along with the corresponding \texttt{<imagine>} description upon determining an image insertion point. The Visualizer then generates the image conditioned on this description, after which the Planner resumes text generation using the full preceding context. This process repeats iteratively until the complete interleaved sequence is produced.

\begin{figure}[ht]
    \centering
    \includegraphics[width=1.0\linewidth]{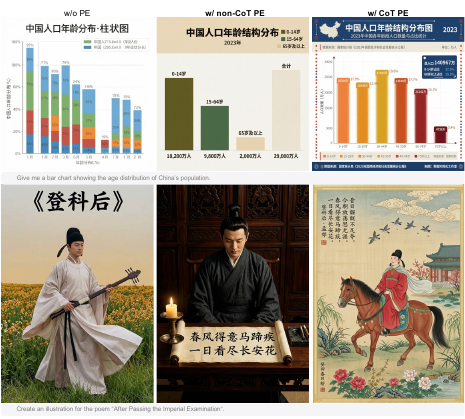}
    \caption{
    Visual comparison of different variants of our PE.
    }
    \label{fig:showcase_pe}
\end{figure}

\subsection{Prompt Enhancer}
Benefiting from the strong image generation capability and excellent image understanding ability, our model can comprehend complex user intent and perform fine-grained layout and compositional design. To fully exploit this advantage, we introduce a {Prompt Enhancer (PE)} that rewrites the user's original input into a structured, detail-rich description, thereby providing more effective guidance for the generation model to produce high-quality images. Specifically, the PE module is responsible for interpreting user intent, enriching scene elements, and performing fine-grained design of the overall image composition and element arrangement, achieving significant improvements shown in Figure~\ref{fig:showcase_pe}.

\paragraph{Model variants.}
To balance generation quality and inference efficiency, we develop two variants of the PE module depending on whether it employs the Chain-of-Thought (CoT) strategy or not:

\begin{itemize}
    \item \textbf{Non-CoT variant.} The non-CoT variant is designed for low-latency scenarios, with inference typically completed within approximately one second. The number of output tokens is constrained to the range of 400--600, enabling efficient inference while maintaining high-quality prompt rewriting. This variant is trained based on Qwen3-VL-2B~\citep{bai2025qwen3vl}. Although the non-CoT model does not produce explicit reasoning traces during inference, reasoning processes are incorporated during the construction of its training data, which helps ensure the quality and reliability of the rewritten outputs.
    
    \item \textbf{CoT variant.} The CoT variant prioritizes generation quality. Inspired by the CoT paradigm, the model generates explicit reasoning steps prior to producing the final rewritten prompt. During inference, the model emulates the reasoning process of a human designer, systematically considering how to enrich scene elements, design the composition, and arrange multiple elements within a coherent spatial layout. The final rewritten prompt is generated based on this reasoning process, providing rich and detailed descriptions of both foreground elements and background context. Consequently, the generated images exhibit strong consistency and richness in both semantic content and spatial structure. The CoT variant typically produces approximately 1,500--2,000 output tokens and is trained based on Qwen3-VL-30B-A3B~\citep{bai2025qwen3vl}.
    
\end{itemize}

\paragraph{Task coverage.} The PE module is a unified model jointly trained on four generation tasks: T2I, I2I, T2S, and TI2S, with task-specific inputs and output formats distinguished via structured prompts.
For T2I tasks, the PE module rewrites the user's textual input into a detail-rich image description with explicit compositional structure. For T2S tasks, the PE module produces a structured list, where each entry corresponds to one image in the output group and follows the same format as the T2I rewrite. The number of entries is determined by the number of images requested by the user.
For reference-guided I2I and TI2S tasks, the PE module produces two key fields: common and difference. The common field describes content that should remain consistent between the input and output images, such as the primary subject or visual style, while the difference field specifies the aspects to be modified. This explicit disentanglement enables the generation model to more precisely identify what should be preserved and what should be altered, thereby facilitating controllable editing with respect to the reference image.

\paragraph{Training data.}
The training data for the PE module are automatically constructed using MLLM models. 
For the CoT variant, each training sample contains both the complete reasoning process and the rewritten result. For the non-CoT variant, although the reasoning process is used during data construction to ensure rewriting quality, the model is trained only on the rewritten result as the learning target.
For both the CoT and non-CoT variants, model training is conducted in two stages: SFT followed by RL.
During the SFT stage, the full dataset is utilized for training a single unified model across all four tasks. Among the training data, T2I accounts for the largest proportion, followed by T2S and TI2S at comparable scales, and I2I comprising a smaller portion. Training samples from all tasks are mixed together for joint training. All training data are curated using the data selection framework developed for DiT SFT, ensuring comprehensive coverage of diverse user requests. During dataset construction, particularly for the T2I and T2S tasks, we incorporate the stylistic strengths of our generation model and guide the rewritten prompts to align with these advantageous styles whenever possible. The primary objective of the SFT stage is to establish the model's fundamental capabilities. Notably, in the T2I task, the PE module serves not only as a prompt rewriting component but also as a prompt classification module. We observe that the underlying generation model already demonstrates strong capabilities across many general tasks. Consequently, prompt rewriting is selectively activated only for specific categories, including text rendering and typography, poster and graphic design, infographics and data visualization, interface and UI design, narrative and multi-panel storyboards, technical diagrams and explanations, and logical reasoning and knowledge presentation.
During the RL stage, we curate a set of high-quality training samples across all tasks for continued joint training of the unified model, characterized by moderate difficulty and improved annotation quality. Specifically, T2I accounts for the largest share, while I2I, T2S, and TI2S are allocated at smaller and comparable scales. These datasets are used to train the model using the GRPO~\citep{shao2024grpo} algorithm. The reward model evaluates the generated outputs across multiple dimensions, including intent fidelity, descriptive richness, and linguistic accuracy.

\begin{figure}[t]
    \centering
    \includegraphics[width=1.0\linewidth]{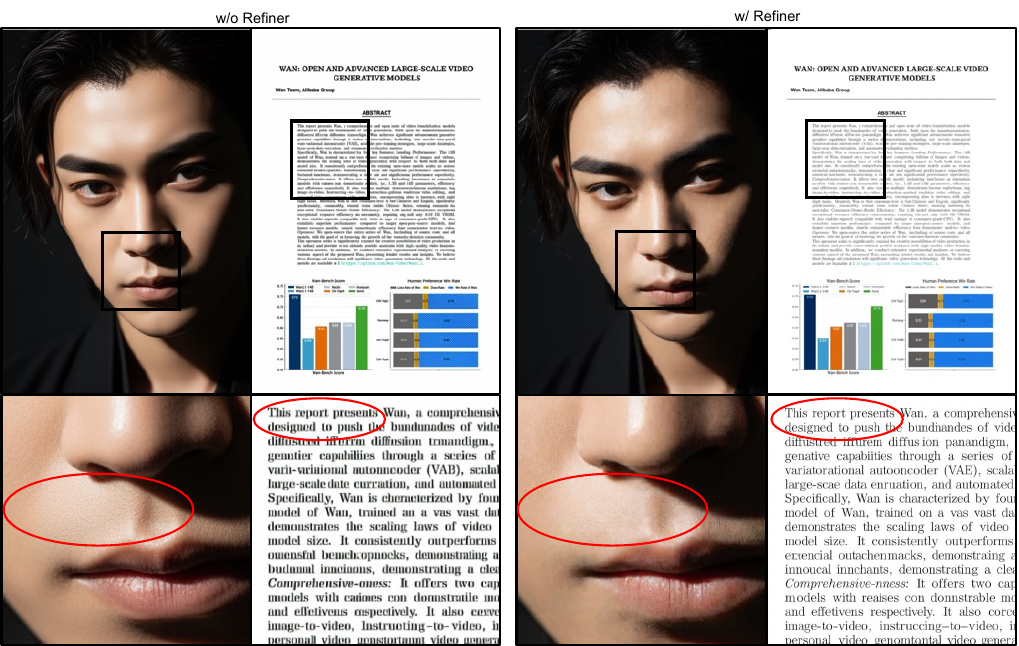}
    \caption{
    Visual comparison of the generated results without the refiner and with the refiner. In high-fidelity mode, the refiner only improves the low-level image details without any structural changes. While in repair mode, refiner will make more significant modifications, such as fixing text errors, eliminating artifacts, \etc.  }
    \label{fig:showcase_refine}
\end{figure}

\subsection{Image Refiner}

\wan supports up to 9 input images and up to 12 output images, allowing the model to more effectively learn and model the relationships across images. However, under a fixed token sequence length, a larger number of input and output images compromise the image resolution. To ensure the quality of the final output, we use an Image Refiner on the outputs of the DiT to enhance output resolution and visual fidelity further. It supports output resolutions of 2K to 4K, delivering enhanced image quality, clarity, and local detail fidelity, as illustrated in Figure~\ref{fig:showcase_refine}.

The refiner features two working modes: repair mode and high-fidelity mode. For images that need deep restoration, it works in repair mode and receives the generated image together with a detailed prompt. In this case, the refiner ensures prompt adherence by refining details and eliminating artifacts, such as misrendered text and image noise. In scenarios where identity preservation is critical, it works in high-fidelity mode because of the low tolerance for identity shift. In this mode, the refiner preserves the integrity of the key visual elements from the input image, ensuring minimal structural changes.
During training, the refiner is trained on images processed through diverse degradation pipelines to handle inputs of varying quality and generation sources. Specifically, clean images are degraded via blurring, noise corruption, and lossy compression. Additionally, a diffusion model is employed to corrupt images via forward noising and subsequent denoising, thereby degrading fine-grained image details.
The refiner is further optimized via distillation and reinforcement learning to improve inference efficiency and visual quality.

\section{Training}

Our training paradigm follows a progressive pipeline that seamlessly integrates semantic comprehension with high-fidelity visual synthesis. We begin with understanding training, establishing a robust cognitive foundation through CT and multi-task SFT, further refined by an on-policy multi-teacher distillation strategy. Empowered by these established semantic priors, generation training translates high-level intent into pixels, advancing from foundational PT to SFT for superior image quality. To ensure outputs resonate with human aesthetics and complex instructions, we incorporate reinforcement learning with human feedback for fine-grained preference alignment. Finally, the collective knowledge is condensed through model distillation, yielding a high-performance and resource-efficient generative framework.

\subsection{Understanding Training}

To achieve a seamless integration of understanding and generation, we optimize the unified understanding–generation paradigm within the MLLM framework through a multi-stage training strategy. Initialized from the Qwen-VL~\citep{bai2025qwen25vltechnicalreport} base model (prior to post-training), our approach comprises three phases: CT to bolster multi-modal understanding, multi-task SFT to generalize generation-oriented tasks, and on-policy multi-teacher distillation to refine real-world performance.

\subsubsection{Understanding Continual Pre-training}

We construct a multi-modal understanding dataset for CT, which incorporates both instruct and thinking samples. Additionally, to mitigate potential degradation in the model's pure-text processing ability, we further build a pure-text CT dataset consisting of a balanced mixture of instruct and thinking samples. During training, this pure-text data is mixed with the multi-modal CT data at a controlled ratio to preserve the model’s textual reasoning and comprehension capabilities. As a result, this mixed-data training strategy not only improves the MLLM’s understanding ability in the instruct mode but also enables a cold start for the thinking mode, thereby equipping the model with stronger deep reasoning capabilities.

\subsubsection{Understanding \& Multi-task Supervised Fine-tuning}

After our model acquires strong multi-modal understanding capabilities, we further refine the unified understanding–generation paradigm by constructing large-scale, high-fidelity text-proxy data, enabling the model to better adapt to user inputs for generation-oriented tasks. The text-proxy data contains dense visual prompts that provide richer image-specific guidance than the relatively sparse surrounding textual context.
To prevent degradation of the model's understanding ability, we jointly train our model using both the text-proxy data and the understanding data. During training, we employ a smaller learning rate than that used for training the pure understanding model, and introduce an item level loss to help the model better distinguish the input and output types across different tasks.
This training strategy offers two key benefits. First, it enables the model to derive precise image generation guidance from purely textual dense prompts, aligning with the text conditioning used by the DiT model during training, thereby facilitating generation through understanding. Second, it equips our model with automatic task inference capability, allowing it to flexibly handle diverse task types within a unified framework, including multi-modal understanding, text-to-image generation, image editing, and interleaved generation.

\subsubsection{On-policy Multi-teacher Distillation}

To enhance our model's performance in real-world scenarios, we construct a query set that represents practical user interaction patterns and generation-task paradigms to further optimize our model. In practical usage, user prompts often contain ambiguous or underspecified intentions, which are far less clear and explicit than those in training data. To tackle this challenge, we employ a high-quality MLLM to perform fine-grained classification of user inputs across multiple dimensions. These resulting classification labels are then used to simulate user intent, constructing a query set that closely mirrors real-world usage patterns. We then train our model on this query set using an on-policy multi-teacher distillation framework, which adaptively selects specialized teacher models for different categories of inputs to transfer the corresponding knowledge. This approach not only further strengthens the model’s understanding capability but also improves the user experience by reducing issues, \eg, confusion arising from ambiguous user intent and unnecessary refusals in generation-oriented tasks.

\begin{table}[t]
\centering
\small 
\caption{Detailed configurations for the training stages of both understanding and generation. U-CT, U\&M SFT, and OPD denote Understanding CT, Understanding \& Multi-task SFT, and On-Policy multi-teacher Distillation, respectively. PT, CT, and SFT denote Pre-training, Continual Pre-training, and Supervised Fine-tuning for the generation part.}
\setlength{\tabcolsep}{2pt} 

\begin{tabularx}{\linewidth}{@{} l *{3}{>{\centering\arraybackslash}X} | *{3}{>{\centering\arraybackslash}X} @{}}
\toprule
 & \multicolumn{3}{c}{\textbf{Understanding}} & \multicolumn{3}{c}{\textbf{Generation}} \\
\textbf{Configuration} & \textbf{U-CT} & \textbf{U\&M SFT} & \textbf{OPD} & \textbf{PT} & \textbf{CT} & \textbf{SFT} \\
\midrule

\multicolumn{7}{@{} l}{\textit{\textbf{Training Process}}} \\ \addlinespace[2pt]
Steps (K)       & 46             & 10               & 0.4              & 713            & 223              & 13               \\
Tokens (T)      & 0.1            & 0.033            & 0.0016           & 13.27          & 8.85             & 0.62             \\
Resolution      & -              & -                & -                & 192/320/640        & 512-2048         & 512-2048         \\
Batch Size (K)  & 1              & 1.6              & 0.256            & 50/23/12       & 5                & 4                \\
\midrule

\multicolumn{7}{@{} l}{\textit{\textbf{Data Distribution}}} \\ \addlinespace[2pt]
Type            & T2T/I2T      & Multi-task*      & -                & T2I/I2I      & \multicolumn{2}{c}{T2I/I2I/T2S/TI2S} \\
Ratio           & 2:7          & 1:1:2:2:4        & -                & 7:3          & \multicolumn{2}{c}{7:2:0.5:0.5} \\
\midrule

\multicolumn{7}{@{} l}{\textit{\textbf{Hyperparameters}}} \\ \addlinespace[2pt]
Optimizer       & AdamW          & AdamW            & AdamW            & \multicolumn{3}{c}{AdamW} \\
Weight Decay    & 0.05           & 0.1              & 0.01             & \multicolumn{3}{c}{0.02} \\
Grad. Norm Clip & 1.0              & 1.0              & 0.2                & \multicolumn{3}{c}{0.5} \\
Uncond. Dropout & -              & -                & -                & \multicolumn{3}{c}{0.05} \\
Learning Rate   & $1 \times 10^{-5}$ & $7 \times 10^{-6}$ & $1 \times 10^{-6}$ & $5 \times 10^{-5}$ & $5 \times 10^{-5}$ & $3 \times 10^{-5}$ \\
EMA Ratio       & -              & 0.9998           & -                & 0.9999         & 0.9999           & 0.9998           \\
\bottomrule
\multicolumn{7}{l}{\scriptsize * Multi-task: T2T / I2T / T2I / I2I / Interleaved generation}
\end{tabularx}
\label{tab:training_details}
\end{table}

\subsection{Generation Training}

Within the unified multi-modal framework, we design three key training stages to develop the model’s image generation capabilities: PT, CT, and SFT. By progressively adjusting the training resolution and data composition across these stages, the model gradually advances from basic image generation to ultra-high-resolution generation.
The detailed configurations are summarized in Table~\ref{tab:training_details}.

\subsubsection{Pre-training}
At this stage, the model primarily learns the fundamental principles of image generation. The training corpus comprises 13.27T tokens, and the model is trained for a total of 713K steps. For generation tasks, we adopt a low-to-medium resolution training strategy, including resolutions of 192, 320, and 640. The training data follows a 7:3 mixture of T2I data and I2I data. The learning rate is set to $5 \times 10^{-5}$, enabling the model to establish a robust visual representation foundation from large-scale image–text pairs.

\subsubsection{Continual Pre-training}
As the model’s capabilities stabilize, the CT stage shifts the training focus toward high-resolution understanding and generation, with training resolutions significantly scaled up, ranging from 512 to 2048. This stage processes 8.85T tokens over 223K training steps. To further enhance cross-modal reasoning and multi-image content generation capabilities, we adjust the data composition by introducing T2S and TI2S tasks, resulting in a mixture ratio of 7:2:0.5:0.5 (T2I : I2I : T2S : TI2S). Through this training strategy, the model learns to generate high-quality visual content under long-context conditions and complex constraints.

\subsubsection{Supervised Fine-tuning}
The SFT stage focuses on enhancing the aesthetic quality of generated images and improving the precision of instruction following. In this stage, the training duration is reduced to $13$K steps (approximately $0.62$T tokens), while the training resolution remains within the high range of 512--2048 to ensure robust performance across different resolutions. To enable more fine-grained parameter updates, the learning rate is reduced to 
$3 \times 10^{-5}$, and the EMA ratio is adjusted to $0.9998$. The training data follows the diversified composition introduced in the CT stage but places greater emphasis on carefully curated subsets, either manually filtered or selected by high-quality models, to ensure that the final generated images achieve superior visual detail and logical consistency.

Across all stages, we employ the AdamW~\citep{loshchilov2017adamw} optimizer with a weight decay of $0.02$ and gradient norm clipping set to $0.5$. In addition, an unconditional dropout rate of $0.05$ is introduced to support classifier-free guidance.

\subsection{Reinforcement Learning with Human Feedback}

\begin{figure}
    \centering
    \includegraphics[width=0.9\linewidth]{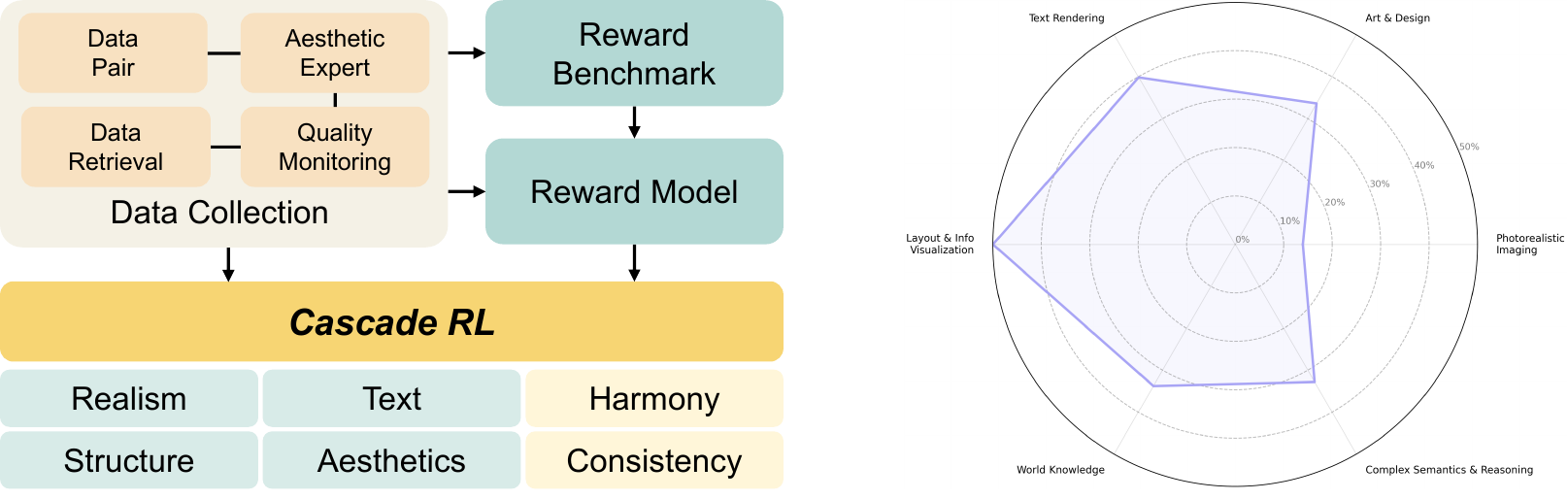}
    \caption{Overview of the reinforcement learning with human feedback framework (Left). Percentage improvement in Win Score of reinforcement learning over the SFT baseline (Right). }
    \label{fig:rl_framework}
\end{figure}

\begin{figure}[!ht]
    \centering
    \includegraphics[width=1.0\linewidth]{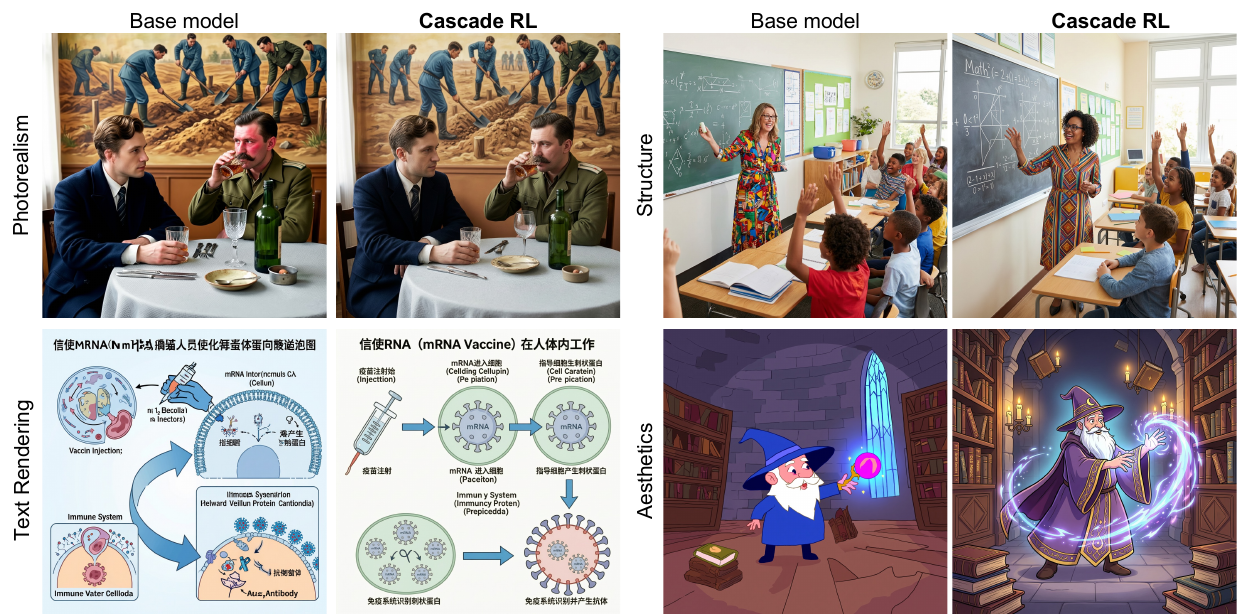}
    \caption{
    Visual comparison of the results generated by the SFT base model and Cascade RL in T2I generation.
    }
    \label{fig:rl_case_t2i}
\end{figure}

\begin{figure}[!ht]
    \centering
    \includegraphics[width=1.0\linewidth]{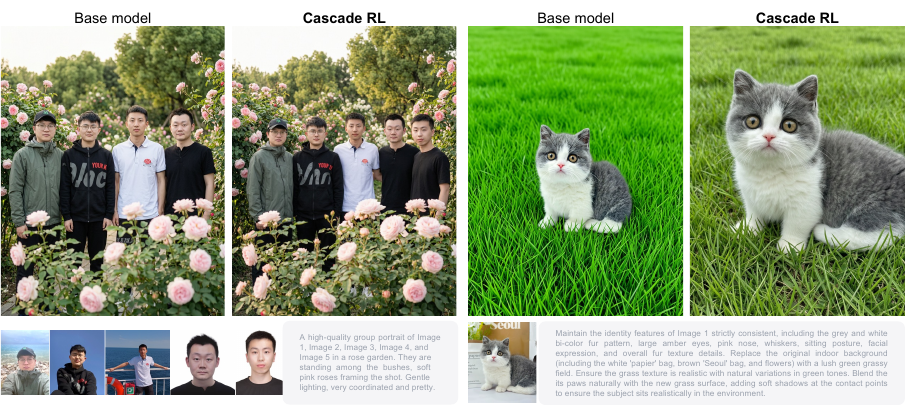}
    \caption{
    Visual comparison of the results generated by the SFT base model and Cascade RL in I2I generation.
    }
    \label{fig:rl_case_i2i}
\end{figure}

After the preceding training stages, the model acquires strong foundational capabilities across multiple dimensions; however, it still struggles to consistently produce high-quality results that meet user expectations. After the SFT stage, we further refine the model by incorporating preference data, reward models (RM), and feedback-driven learning algorithms, as illustrated in Figure~\ref{fig:rl_framework} (left). Building on these components, we introduce a cascaded domain-wise reinforcement learning framework (\textbf{Cascade RL}) to adjust the model's output preferences, thereby closely aligning the generated results with human preferences.

\subsubsection{Data}
During the construction of the reinforcement learning dataset, we ensure that the prompts in the SFT dataset and the reinforcement learning dataset are mutually exclusive, thereby preventing the model from achieving high rewards by recalling responses memorized during SFT. In addition, we employ an MLLM to filter and categorize large-scale data, facilitating the subsequent construction of domain-wise datasets. Finally, we establish a comprehensive pipeline composed of data pair construction, aesthetic expert training and annotation, quality monitoring, and data retrieval, enabling the rapid accumulation of high-quality data while maintaining strict quality control.

\subsubsection{Reward Model}
Typically, annotated data are divided into training and validation sets to train and evaluate the RM used during the reinforcement learning process. However, in practice, we observe that this approach often leads to a preference overfitting issue, where the held-out validation split may not reliably reflect the reward model's effectiveness in guiding RL.
To address this issue, we construct a reward benchmark composed of more diverse data sources to obtain reward models with improved consistency and generalization. Based on the benchmark, we conduct a series of ablation studies to determine the appropriate backbone model configuration, including backbone architecture design, model scaling, data scaling, and the choice of loss functions.

\subsubsection{Training Recipe}
After the aforementioned steps, we introduce a comprehensive reinforcement learning framework termed Cascade RL. This framework integrates multiple RL optimization methods, including Direct Preference Optimization (DPO)~\citep{rafailov2023direct}, ReFL~\citep{xu2023imagereward}, DenseGRPO~\citep{deng2026densegrpo}, and in-house reinforcement learning with flow matching anchors (ReFMA). The ReFMA method introduces an anchor-based control mechanism that effectively balances optimization objectives across different noise levels while leveraging gradient signals from the reward model to guide policy optimization.
By leveraging the strengths of these algorithms, the framework employs a cascaded training strategy to progressively enhance different dimensions of the model's capabilities. This design enables controllable preference alignment and facilitates the efficient extraction of fine-grained signals for optimizing more subjective attributes of the model. Throughout the entire Cascade RL process, the model exhibits substantial improvements in photorealism, structural correctness, text rendering, and overall aesthetic quality. Meanwhile, in the image editing task, the consistency and harmony of the output results have also been improved.

\subsubsection{Results}

In various scenarios, the winning rate score of the proposed Cascade RL improved by a range of \textbf{15\% to 50\% } compared to the SFT baseline (Figure~\ref{fig:rl_framework}, right). 
As shown in Figure~\ref{fig:rl_case_t2i}, the Cascade RL plays a crucial role in enhancing the overall performance of our diffusion model, achieving significant improvements in photorealism, structural correctness, text rendering, and overall aesthetic quality. 
The visual comparison in Figure~\ref{fig:rl_case_i2i} shows that the consistency and harmony of the results generated by Cascade RL for the I2I tasks have also been improved. In Figure~\ref{fig:rl_morecase}, more visual comparisons are presented.

\subsection{Model Distillation}

\begin{figure}[t]
    \centering
    \includegraphics[width=1.0\linewidth]{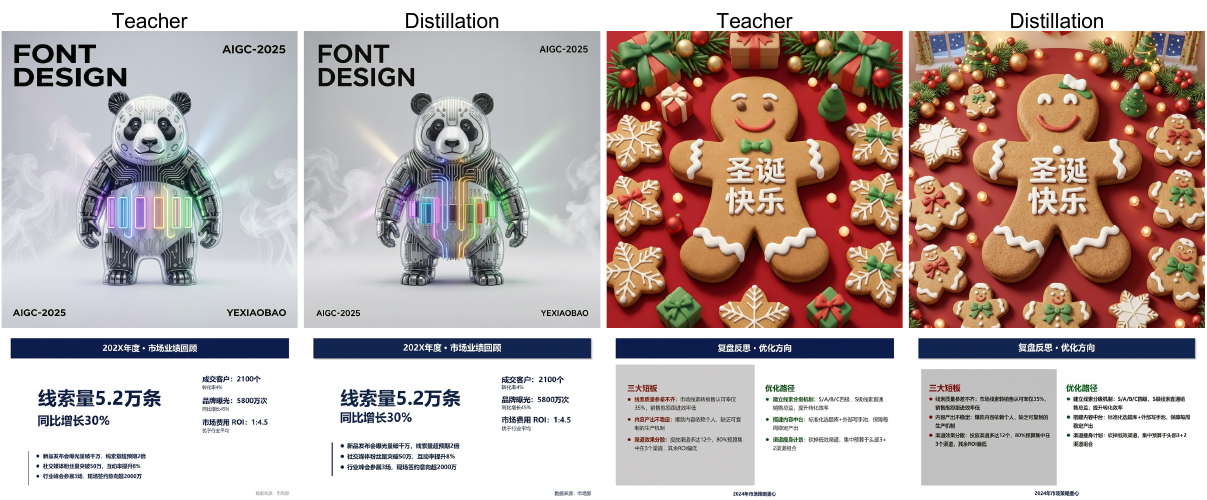}
    \caption{
    Visual comparison of the results generated by the teacher model and the distilled student model.
    }
    \label{fig:distill}
\end{figure}

The primary objective of this stage is inference step distillation, aimed at substantially reducing the base model’s latency to meet the stringent efficiency demands of real-world applications. Despite the efficiency gains afforded by its moderate parameter scale, the base model’s original iterative generation scheme and classifier-free guidance (CFG)~\citep{ho2022cfg} continue to impose a substantial computational burden. The resulting high number of function evaluations (NFE) requirement remains a critical bottleneck for practical deployment, creating an inherent trade-off between preserving generation quality and meeting the low-latency demands of interactive applications. While recent step-distillation methods accelerate sampling, they tend to exhibit degradation in generative fidelity, particularly in preserving high-frequency facial details and text rendering capability. This gap is particularly noticeable in high-capacity models, where the teacher’s complex denoising dynamics are harder to replicate with fewer steps. To address this challenge, we train a distilled model that closely approximates the teacher’s denoising trajectory with significantly fewer sampling steps. Specifically, we introduce a multi-stage progressive distillation framework that integrates two complementary paradigms: trajectory distillation and distribution matching distillation~\citep{lu2024simplifying,luo2025learning}. This framework strikes a balance during generation by preserving global structural coherence, retaining high-frequency visual details, and improving text alignment fidelity, as illustrated in Figure~\ref{fig:distill}. Consequently, our compression of the inference process yields a speedup exceeding $8 \times$, while preserving generation fidelity.

\section{Results}

\subsection{Understanding Results}

\newcommand{\rot}[2][45]{\makebox[1em][l]{\rotatebox{#1}{#2}}}

As a unified model, we first evaluate its performance on understanding-relevant benchmarks. Since our model is built upon Qwen-VL, we primarily compare with models based on the comparable-scale LLM or VL backbone. The comparison covers both understanding-only and unified models across Image Understanding and Video Understanding benchmarks. The results demonstrate that our model significantly outperforms previous unified models and achieves or surpasses the performance levels of dedicated understanding-only models. For Image Understanding, we selected 7 classical multi-modal benchmarks to assess the model's performance, including MMMU~\citep{yue2024mmmu}, MMStar~\citep{chen2024mmstar}, MathVista~\citep{lu2023mathvista}, HalluBench~\citep{guan2024hallusionbench}, MMBench~\citep{liu2024mmbench}, OCRBench~\citep{fu2024ocrbench}, and AI2D~\citep{hiippala2021ai2d}. The results shown in Table~\ref{tab:understanding_results} demonstrate that our model exhibits superior instruct and thinking capabilities: in the instruct mode, it achieves an average score 1.6 points higher than the baseline and remains comparable to other open-source understanding-only instruct models. Concurrently, in thinking mode, our model outperforms the baseline by 5.4 points and exceeds the performance of other open-source understanding-only thinking models. Compared to other unified models, our approach maintains a distinct and absolute lead. Regarding Text Understanding, our model demonstrates significant improvements over the showcased understanding-only models in both instruct and thinking modes on four common benchmarks: AIME~\citep{patel2024aime}, GPQA~\citep{rein2024gpqa}, HLE~\citep{phan2025hle}, and LCBV6~\citep{jain2024lcbv}.

\begin{table}[htbp]
\centering
\caption{
Comparison of model performance on representative understanding metrics. ``--'' indicates that the indicator is missing in the relevant report.}

\small
\setlength{\tabcolsep}{3pt}
\begin{tabular}{llccccccccccccc}
\toprule
 & & \multicolumn{8}{c}{Image Understanding} & \multicolumn{5}{c}{Text Understanding} \\
\cmidrule(lr){3-10} \cmidrule(lr){11-15}
Type & Model & \rot{MMMU} & \rot{MMStar} & \rot{MathVista} & \rot{HalluBench} & \rot{MMBench} & \rot{OCRBench} & \rot{AI2D} & \rot{Avg.} &\rot{AIME2025} & \rot{GPQA} & \rot{HLE} & \rot{LCBV6} & \rot{Avg.}\\
\midrule
\multirow{7}{*}{Und. Only} 
 & Qwen2-VL-7B & 54.1 & 60.7 & 58.2 & 50.6 & 80.7 & 86.6 & 83.0 & 67.7& -- & 30.3 & 2.8 & 9.7 &10.7\\
 & Qwen2.5-VL-7B & 58.6 & 63.9 & 68.2 & 52.9 & 82.6 & 86.4 & 83.9 &70.9 & -- & 34.3 & 2.7 & 14.9 & 13.0\\
 & Ovis2-8B & 57.4 & 64.6 & 71.8 & 56.3 & 83.6 & 89.1 & 86.6 &72.8 & -- & 35.9 & 4.2 & 20.0 & 15.0\\
 & InternVL3-8B & 62.7 & 68.2 & 71.6 & 49.9 & 81.7 & 88.0 & 85.2 &72.5& -- & 36.4 & 4.4 & 16.5 & 14.3\\
 & InternVL2.5-8B & 56.2 & 62.8 & 64.5 & 50.1 & 83.2 & 82.2 & 84.5 &69.1& 3.3 & 28.3 & 3.3 & 10.3 & 11.3\\
 & Kimi-VL-2506 (Thinking) & 64.0 & 70.4 & 80.1 & 59.8 & 84.4 & 86.9 & 81.9 &75.4& 33.3 & 45.5 & 2.3 & 29.7 & 27.7\\
\midrule
\multirow{8}{*}{Unified} 
 & BAGEL-7B & 55.3 & -- & 73.1 & -- & 85.0 & -- & -- & -- & -- & -- &-- & -- &--\\
 & UniWorld-V1 & 58.6 & -- & -- & -- & 83.5 & -- & -- & -- & -- & -- &-- & -- &--\\
 & MUSE-VL & 50.1 & -- & 55.9 & -- & 81.8 & -- & -- & -- & -- & -- &-- & --& --\\
 & Emu3 & 31.6 & -- & -- & -- & 58.5 & 68.7 & 70.0 & -- & -- & -- &-- & -- &--\\
 & MetaMorph & 41.8 & -- & -- & -- & 75.2 & -- & -- & -- & -- & -- &-- & -- &--\\
 & Janus-Pro-7B & 41.0 & -- & -- & -- & 79.2 & -- & -- & -- & -- &-- & -- & -- &--\\
 & BLIP3-o 8B & 50.6 & -- & -- & -- & 83.5 & -- & -- & -- & -- &-- & -- & -- &--\\
 \rowcolor{Gray}
 & Ours Instruct & 58.0 & 68.0 & 73.5 & 55.4 & 82.7 & 84.8 & 85.1 &72.5& 30.0 & 44.8 & 3.4 & 32.6 &27.7\\
 \rowcolor{Gray}
 & Ours Thinking & 67.4 & 72.8 & 82.3 & 57.7 & 83.5 & 84.1 & 86.1 &76.3& 33.3 & 47.0 & 3.9 & 31.4 &28.9\\
\bottomrule
\end{tabular}
\label{tab:understanding_results}
\end{table}

\subsection{Generation Results}
\label{sec:gen_result}

In this section, we provide a comprehensive evaluation of our proposed model through both quantitative analysis and qualitative visualizations across various generative tasks and stages.

\subsubsection{Quantitative Evaluation}
To evaluate the comprehensive performance of our model against existing state-of-the-art methods, we conduct a multi-dimensional quantitative analysis. As illustrated in the radar chart in Figure~\ref{fig:radar_report}, our model demonstrates superior capabilities across several key benchmarks.

\subsubsection{Qualitative Results}
\begin{figure}[!ht]
    \centering
    \includegraphics[width=0.98\textwidth]{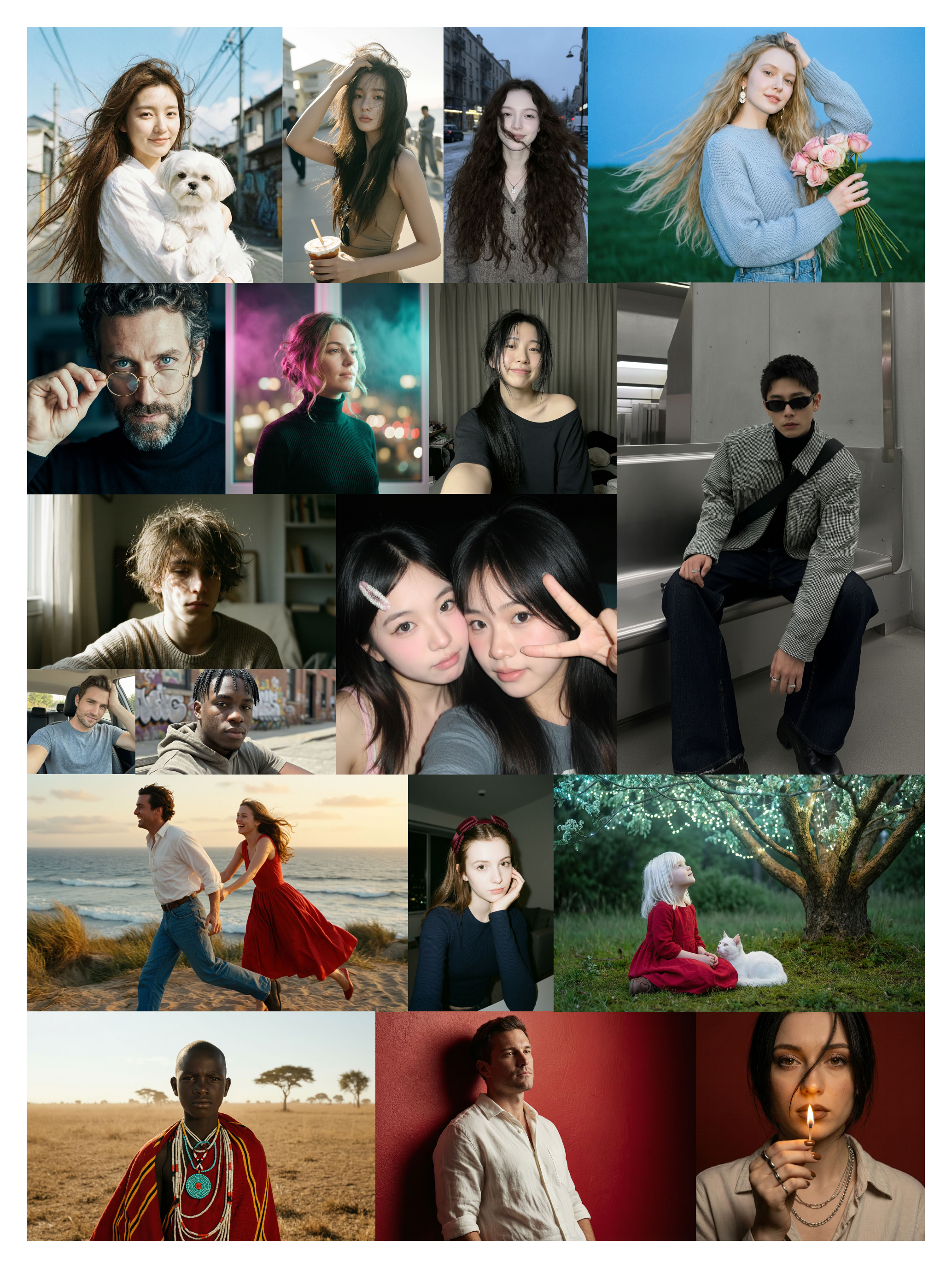}
    \caption{More visual demonstration of diverse capabilities in T2I generation.}
    \label{fig:showcase_t2i_more_2}
\end{figure}

\begin{figure}[!ht]
    \centering
    \includegraphics[width=0.98\textwidth]{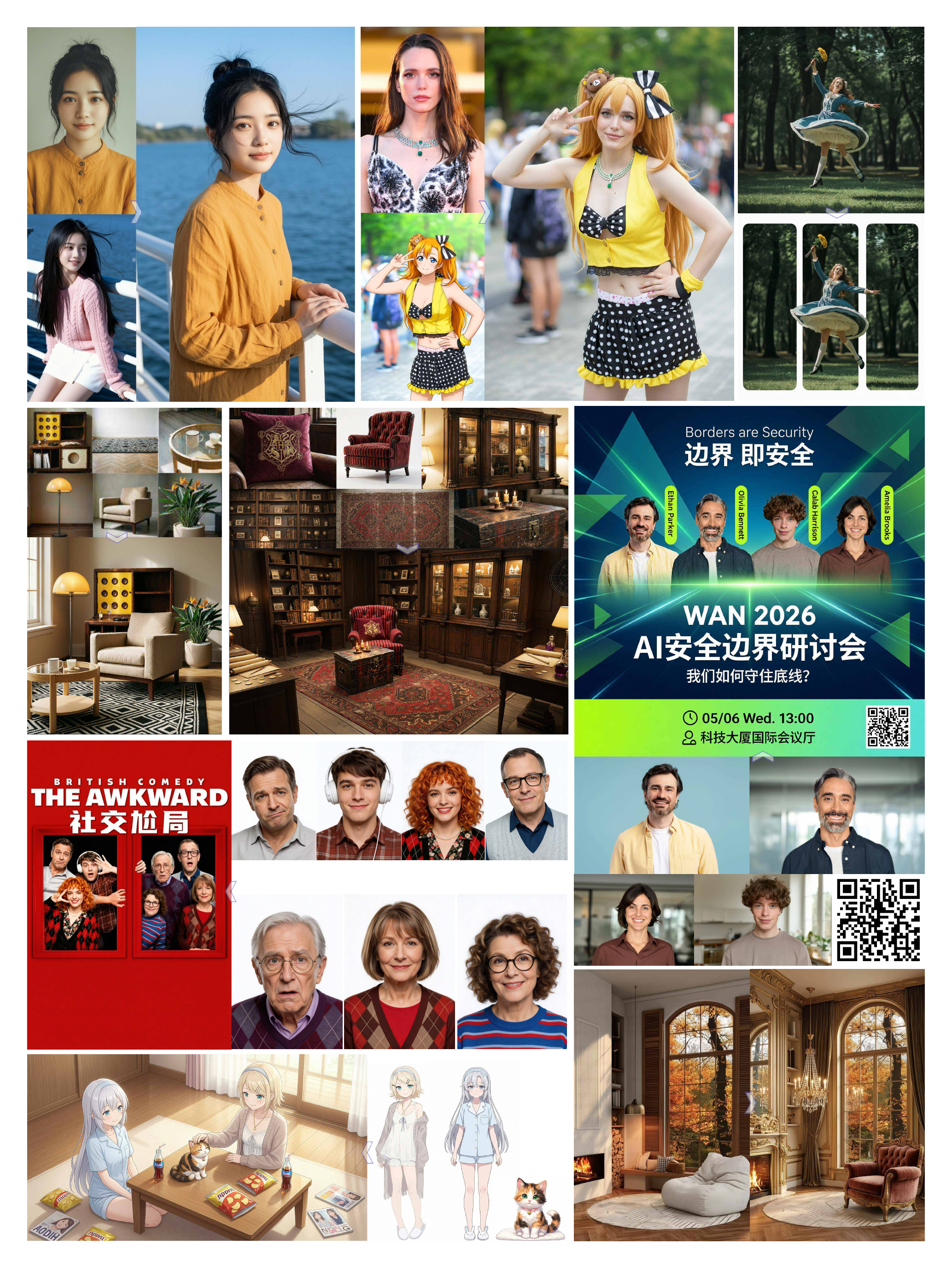}
    \caption{More visual demonstration of diverse capabilities in I2I generation.}
    \label{fig:showcase_i2i_more_1}
\end{figure}

\begin{figure}[!ht]
    \centering
    \includegraphics[width=0.98\textwidth]{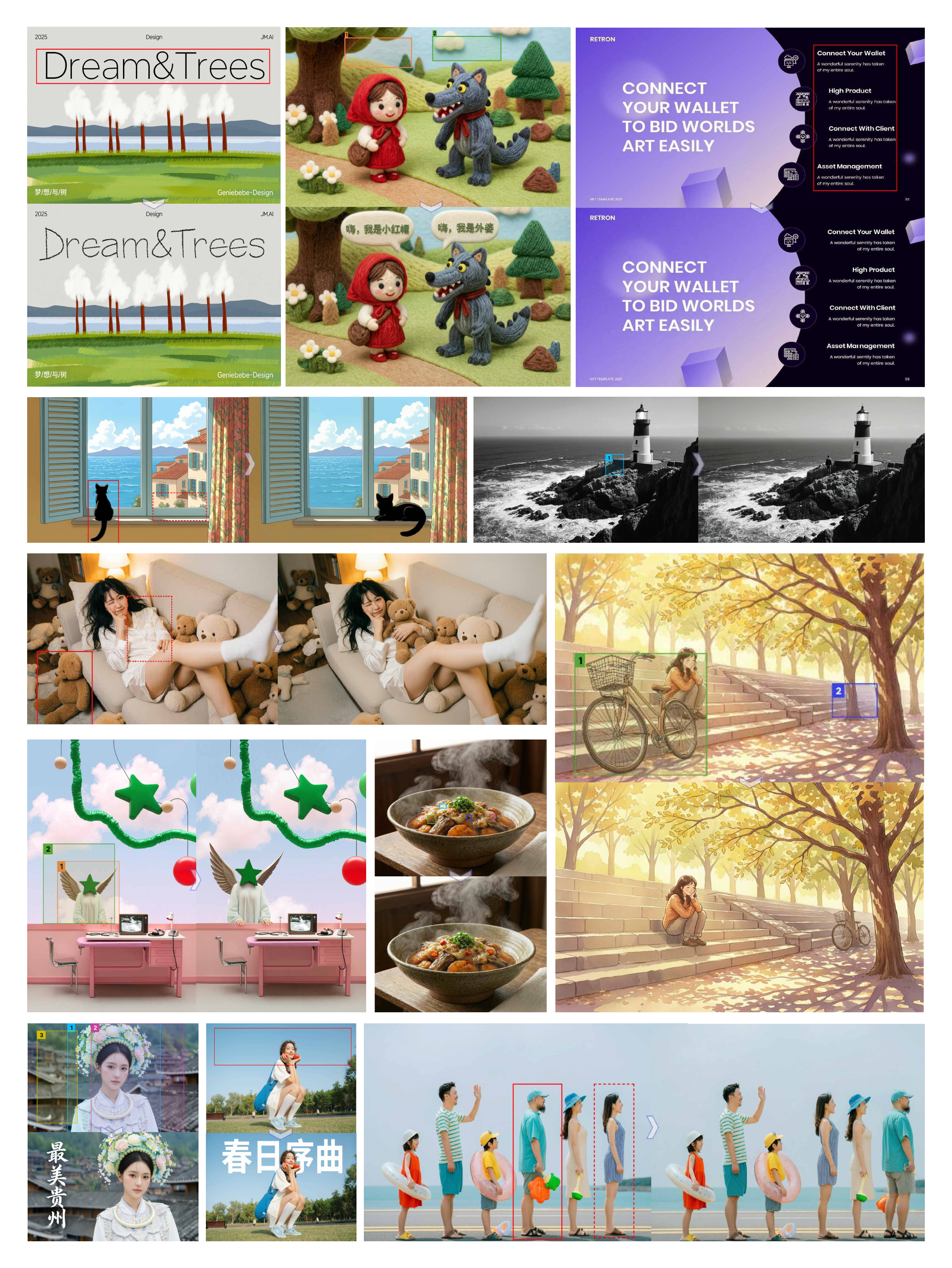}
    \caption{More visual demonstration of diverse capabilities in the interactive editing generation.}
    \label{fig:showcase_i2i_more_2}
\end{figure}

\begin{figure}[!ht]
    \centering
    \includegraphics[width=0.98\textwidth]{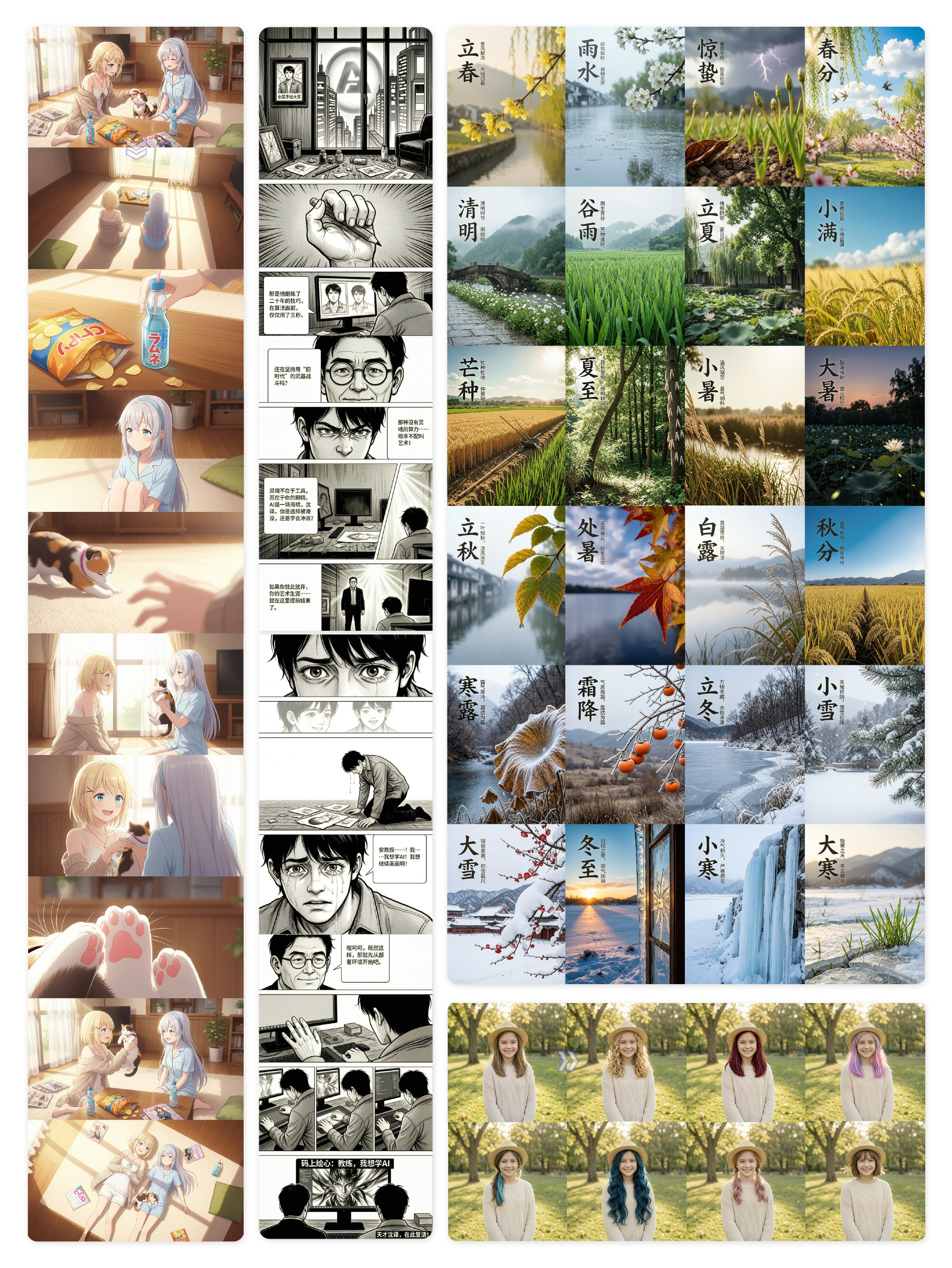}
    \caption{More visual demonstration of diverse capabilities in T2S and TI2S generation.}
    \label{fig:showcase_t2s_more_1}
\end{figure}

\begin{figure}[!ht]
    \centering
    \includegraphics[width=0.98\textwidth]{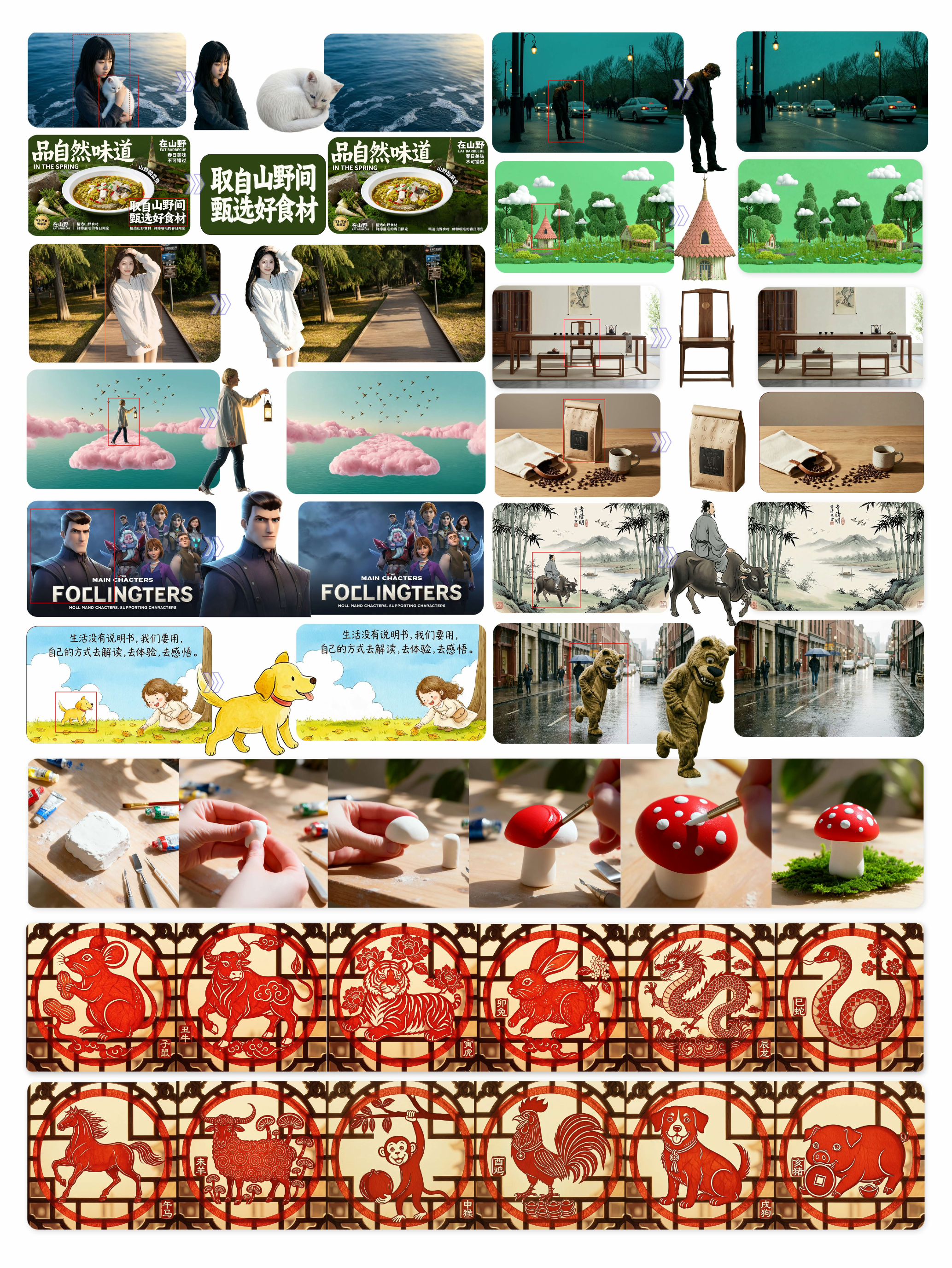}
    \caption{More visual demonstration of diverse capabilities in T2S and TI2S generation.}
    \label{fig:showcase_t2s_more_2}
\end{figure}

\begin{figure}[!ht]
    \centering
    \includegraphics[width=0.98\textwidth]{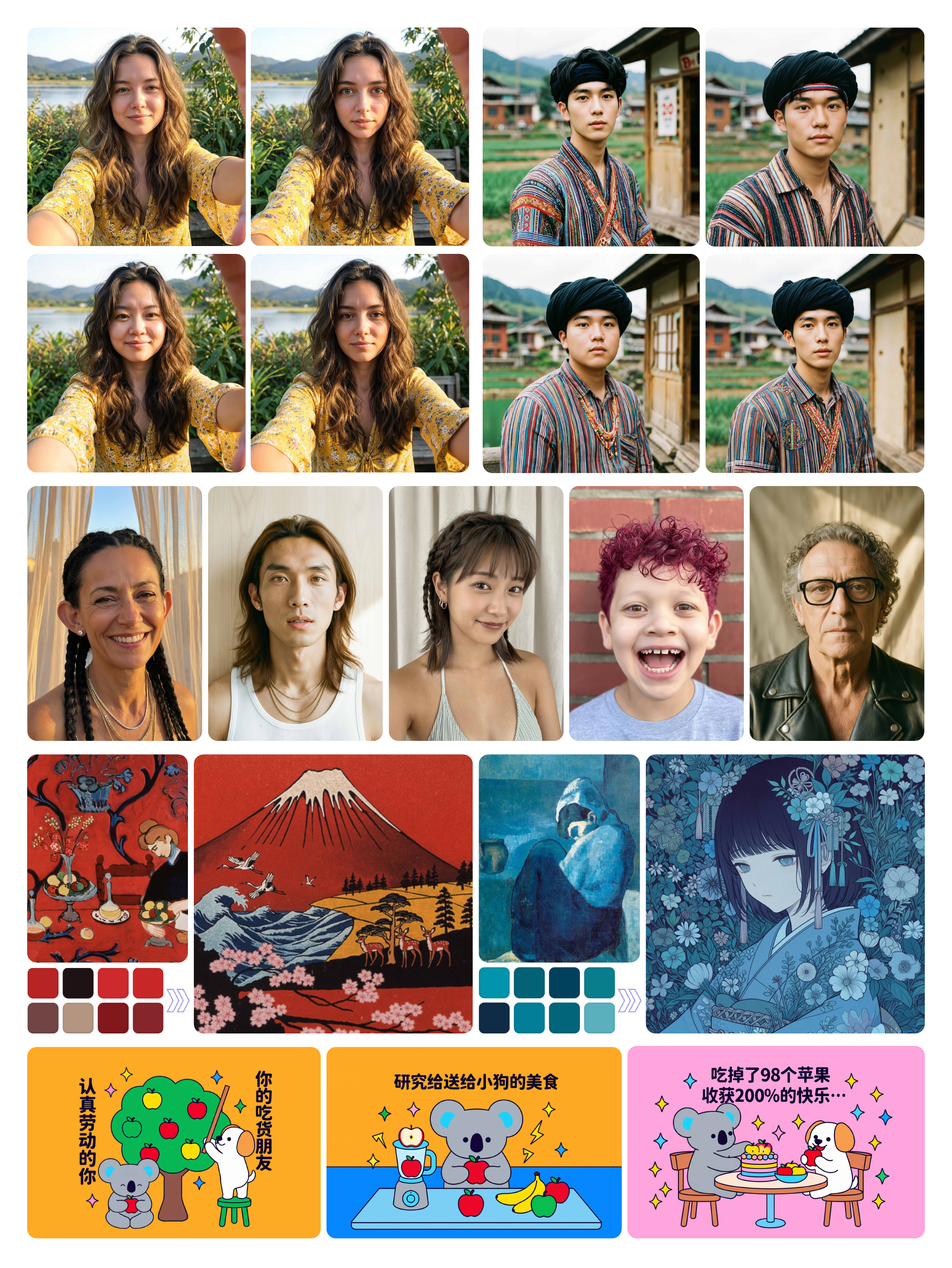}
    \caption{More visual demonstration of diverse capabilities in hyper-diverse realistic portrait generation, palette-guided generation, logical image series generation, \etc.}
    \label{fig:showcase_mix_more_1}
\end{figure}

\begin{figure}[!ht]
    \centering
    \includegraphics[width=0.98\textwidth]{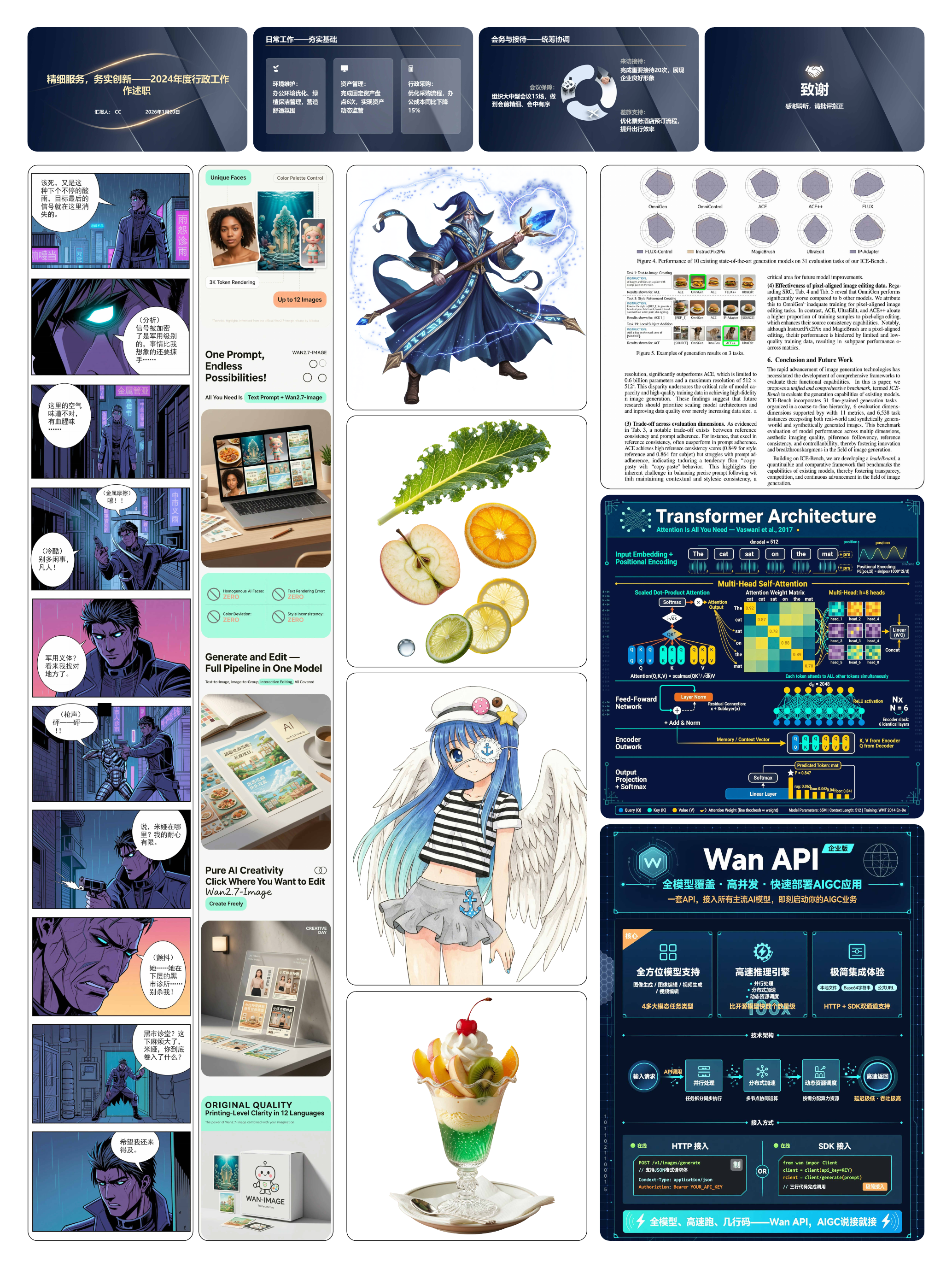}
    \caption{More visual demonstration of diverse capabilities in logical image series generation, extreme aspect ratio adaptation, true alpha-channel generation, ultra-long text rendering, \etc.}
    \label{fig:showcase_mix_more_2}
\end{figure}

\begin{figure}[!ht]
    \centering
    \includegraphics[width=0.98\textwidth]{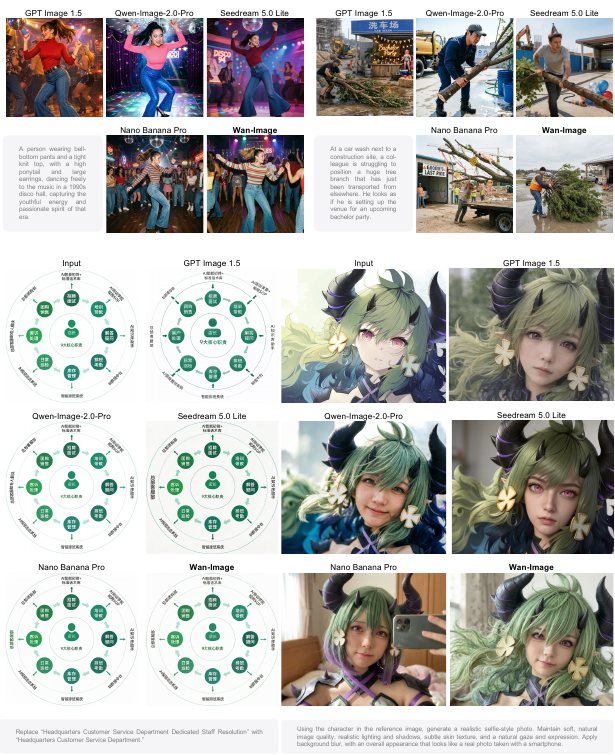}
    \caption{Visual comparison with several models in T2I and I2I generation.}
    \label{fig:compare_t2i_i2i}
\end{figure}

\begin{figure}
    \centering
    \includegraphics[width=0.87\textwidth]{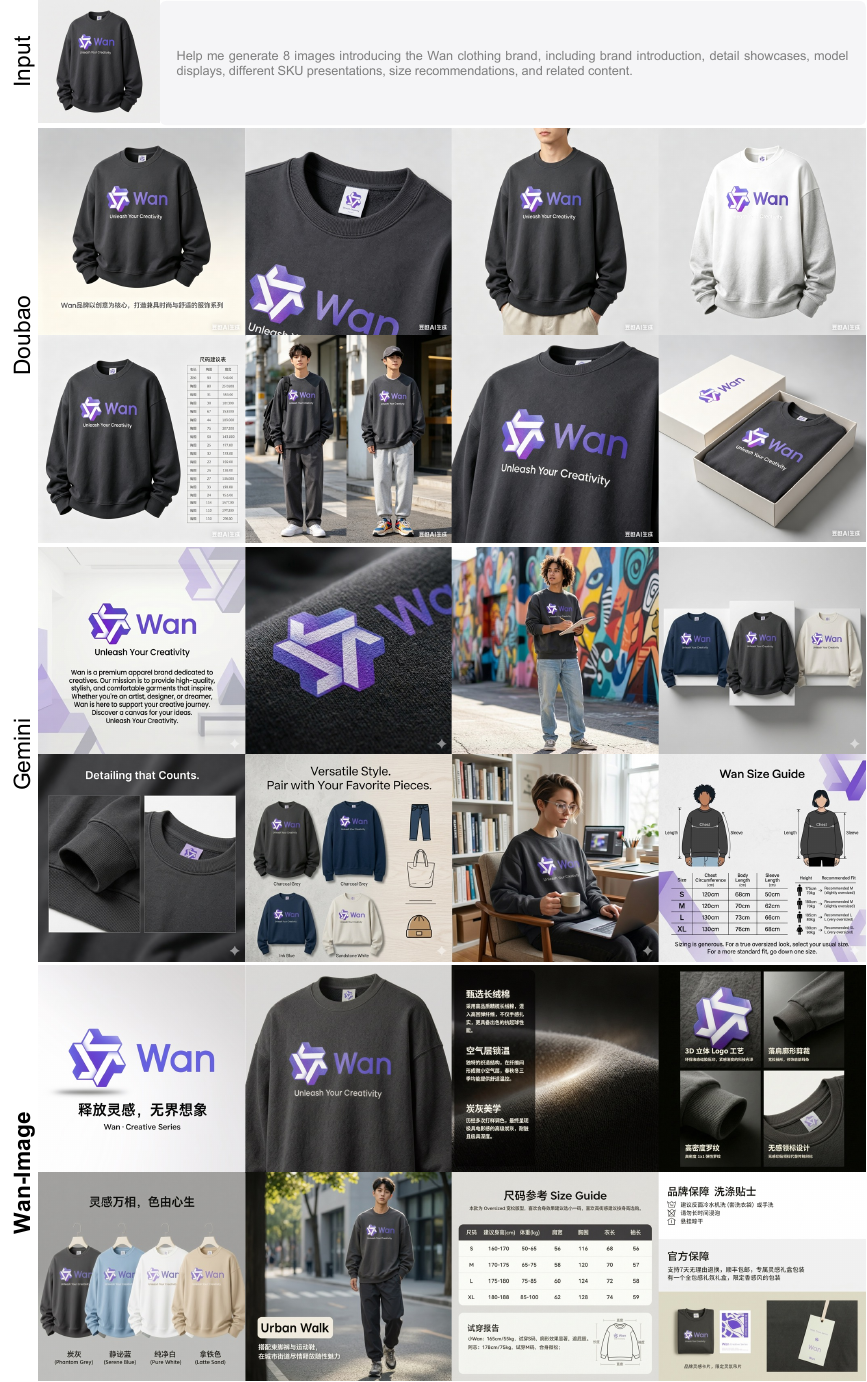}
    \caption{Visual comparison with several models in TI2S generation.}
    \label{fig:compare_ti2g}
\end{figure}

We demonstrate the versatility of our model across four primary categories: Text-to-Image, Image-to-Image, Text-to-Image-Series, and Text-Image-to-Image-Series. Representative results are visualized in Figure~\ref{fig:showcase_t2i_more_2}, \ref{fig:showcase_i2i_more_1}, \ref{fig:showcase_i2i_more_2}, \ref{fig:showcase_t2s_more_1}, \ref{fig:showcase_t2s_more_2}, \ref{fig:showcase_mix_more_1}, and~\ref{fig:showcase_mix_more_2}. Additionally, we make a comparison with several models in Figure~\ref{fig:compare_t2i_i2i} and~\ref{fig:compare_ti2g}, including GPT Image 1.5~\citep{openai2025gptimage1p5}, Qwen-Image-2.0-Pro~\citep{wu2025qwen}, Seedream 5.0 Lite~\citep{seedream2025seedream}, Nano Banana Pro~\citep{google2025nanobananapro}, Doubao~\citep{doubao2024}, and Gemini~\citep{team2023gemini}.

\paragraph{Text-to-Image.} 
Our model generates high-fidelity images with rigorous adherence to complex prompts. Beyond standard generation, it supports advanced features including ultra-long text rendering, extreme aspect ratios, hyper-diverse portrait, color palette-guided generation, and alpha-channel generation, demonstrating superior flexibility in specialized creative workflows (see Figure~\ref{fig:showcase_t2i_more_2}, \ref{fig:showcase_mix_more_1}, \ref{fig:showcase_mix_more_2}, and~\ref{fig:compare_t2i_i2i}).

\paragraph{Image-to-Image.}
The model supports a wide array of tasks, such as instruction-based editing, image-based referencing, and interactive manipulation. It successfully preserves original layouts and maintains robust identity consistency while accurately executing target modifications (see Figure~\ref{fig:showcase_i2i_more_1}, \ref{fig:showcase_i2i_more_2}, and~\ref{fig:compare_t2i_i2i}).

\paragraph{Text-to-Image-Series and Text-Image-to-Image-Series.}
A core strength of our model is its ability to generate coherent image sequences. Given a textual narrative, the model generates a sequence of images in a single pass that maintains strong logical progression and visual continuity. By leveraging both textual instructions and initial image prompts, the model produces a series that exhibits high fidelity to the reference style and identity, ensuring seamless temporal consistency across the entire sequence (see Figure~\ref{fig:showcase_t2s_more_1}, \ref{fig:showcase_t2s_more_2}, and~\ref{fig:compare_ti2g}).

\subsubsection{Analysis of Different Phases}
To better understand the model's evolution, we evaluate how successive training phases, reinforcement learning, and the prompt enhancer collectively shape the final results. This systematic comparison highlights the specific role each stage plays in achieving high-quality and human-aligned generation.

\begin{figure}
    \centering
    \includegraphics[width=0.92\textwidth]{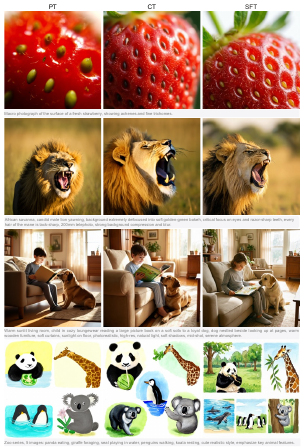}
    \caption{Visual comparison of the generated results across different training stages.}
    \label{fig:showcase_train_stage}
\end{figure}

\paragraph{Evolution of main training stages.} 
As illustrated in Figure~\ref{fig:showcase_train_stage}, we compare results across the Pre-training, Continual Pre-training, and Supervised Fine-tuning stages. The transition from PT to SFT shows a clear trajectory of improved instruction following and localized detail refinement.

\begin{figure}
    \centering
    \includegraphics[width=1.0\linewidth]{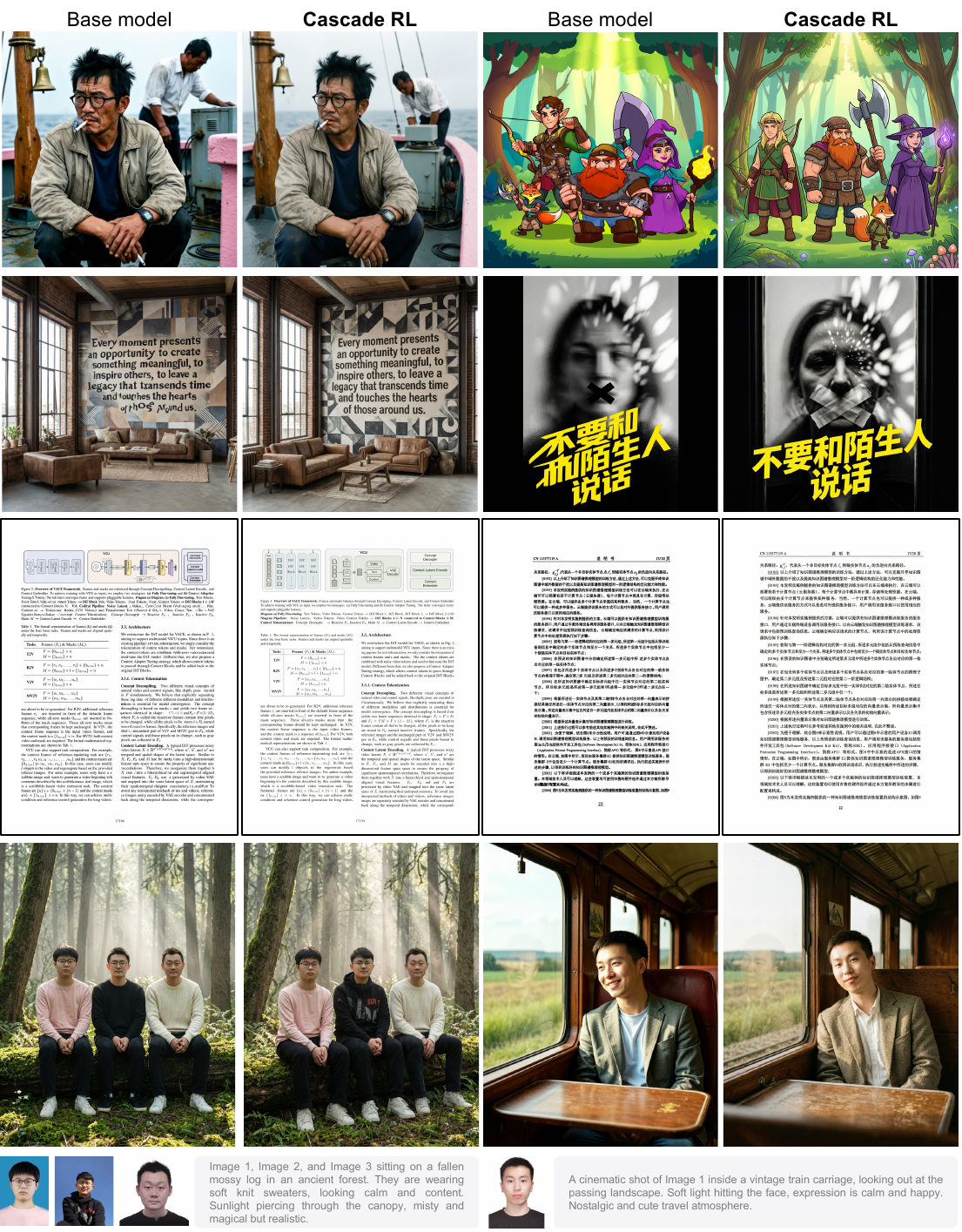}
    \caption{
    Visual comparison of the results generated by the SFT base model and Cascade RL in T2I and I2I tasks.
    }
    \label{fig:rl_morecase}
\end{figure}

\paragraph{Impact of reinforcement learning.} 
The RL phase focuses on aligning the model with human aesthetic preferences. While pre-RL outputs may occasionally suffer from anatomical distortions or unnatural compositions, the post-RL model produces significantly more `natural' and visually appealing results with improved global harmony (see Figure~\ref{fig:rl_morecase}).

\begin{figure}
    \centering
    \includegraphics[width=0.95\linewidth]{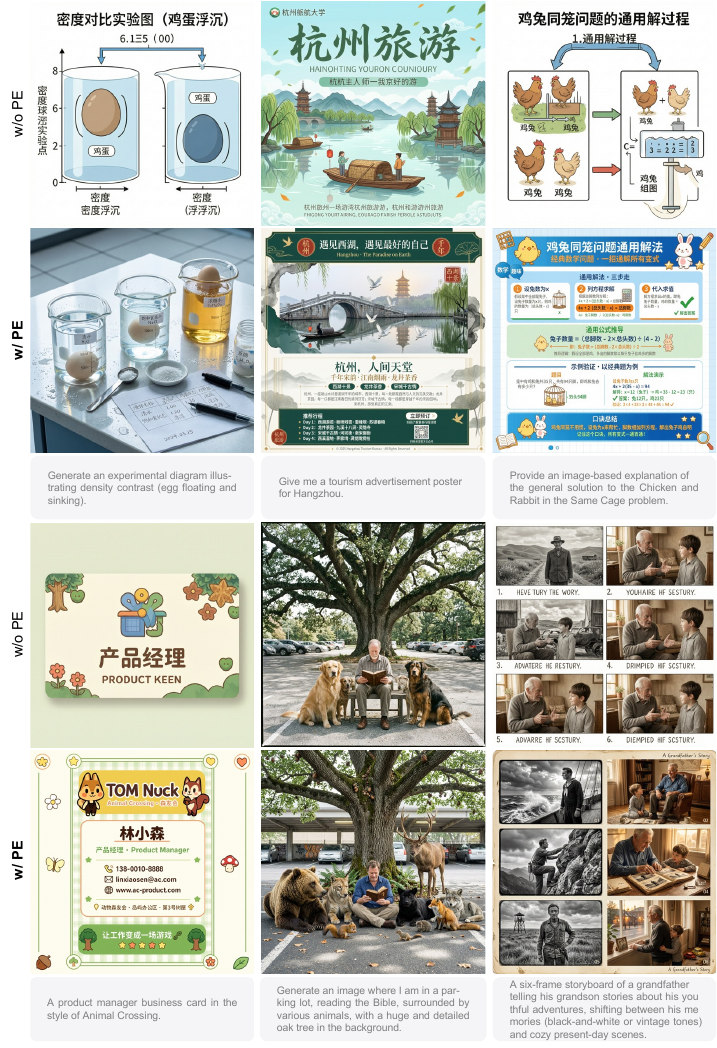}
    \caption{
    Visual comparison of the results generated without the PE and with the PE.
    }
    \label{fig:morecase_pe}
\end{figure}

\paragraph{Impact of prompt enhancer.} 
We evaluate the efficacy of the PE module by comparing raw user inputs with PE-refined prompts. The enhanced versions exhibit significantly richer textures, more sophisticated lighting, and complex artistic compositions without deviating from the user's original semantic intent (see Figure~\ref{fig:morecase_pe}).

\section{Conclusion}
In this report, we present \wan, a unified and high-performance visual generation system engineered to serve as a next-generation professional productivity tool. By conceptually synergizing an MLLM-based Planner for deep semantic reasoning with a DiT-based Visualizer, \wan establishes a robust framework that transcends the limitations of casual image synthesis. Our system demonstrates unprecedented versatility across a wide spectrum of professional tasks, enabling ultra-long text rendering with complex typography, extreme aspect ratio adaptation up to 1:8, and high-efficiency 4K generation without compromising speed.

Through meticulous data scaling and multi-task optimization, \wan empowers creators with fine-grained control, ranging from palette-guided generation via explicit hex-code proportions to hyper-diverse realistic portrait steering and multi-subject identity-preserving generation. Furthermore, Wan fulfills rigorous industrial requirements by supporting the creation of logical image series with thematic consistency, multi-modal interactive editing with precise localized control, and true alpha-channel generation for immediate design integration. Extensive evaluations show that \wan achieves highly competitive performance against state-of-the-art models, fulfilling the complex demands of real-world design workflows. As a comprehensive and intelligent assistant, \wan represents a significant step toward bridging the gap between open-ended creative exploration and professional-grade content production.
\section{Authors}

\subsection[Core Contributors]{Core Contributors\footnote{Core Contributors are listed in alphabetical order of the first name.}}
Chaojie Mao, Chen-Wei Xie, Chongyang Zhong, Haoyou Deng, Jiaxing Zhao, Jie Xiao, Jinbo Xing, Jingfeng Zhang, Jingren Zhou, Jingyi Zhang, Jun Dan, Kai Zhu, Kang Zhao, Keyu Yan, Minghui Chen, Pandeng Li, Shuangle Chen, Tong Shen, Yu Liu, Yue Jiang, Yulin Pan, Yuxiang Tuo, Zeyinzi Jiang, Zhen Han

\subsection[Contributors]{Contributors\footnote{Contributors are listed in alphabetical order of the first name.}}
Ang Wang, Bang Zhang, Baole Ai, Bin Wen, Boang Feng, Feiwu Yu, Gang Wang, Haiming Zhao, He Kang, Jianjing Xiang, Jianyuan Zeng, Jinkai Wang, Junjie Zhou, Ke Sun, Linqian Wu, Pei Gong, Pingyu Wu, Ruiwen Wu, Tongtong Su, Wenmeng Zhou, Wenting Shen, Wenyuan Yu, Xianjun Xu, Xiaoming Huang, Xiejie Shen, Xin Xu, Yan Kou, Yangyu Lv, Yifan Zhai, Yitong Huang, Yun Zheng, Yuntao Hong, Zhe Zhang, Zhicheng Zhang


\newpage

\clearpage

\bibliographystyle{assets/plainnat}
\bibliography{main}


\end{document}